\documentclass[sigconf]{acmart}
\AtBeginDocument{%
  }

\usepackage{xpatch}
\usepackage{adjustbox}
\usepackage{multirow}
\usepackage{boldline}
\usepackage{wrapfig}
\usepackage{array}
\usepackage{multirow}
\usepackage{makecell}
\usepackage{tabularray}
\usepackage{algorithmic}
\usepackage{tikz}
\usepackage{xcolor} 
\usepackage[misc]{ifsym}

\setcopyright{acmlicensed}
\copyrightyear{2026}
\acmYear{2026}
\setcopyright{cc}
\setcctype{by-nc-nd}
\acmConference[KDD'26] {Proceedings of the 32nd ACM SIGKDD Conference on Knowledge Discovery and Data Mining V.2}{August 9--13, 2026}{Jeju Island, Republic of Korea.}
\acmBooktitle{Proceedings of the 32nd ACM SIGKDD Conference on Knowledge Discovery and Data Mining V.2 (KDD 2026), August 9--13, 2026, Jeju Island, Republic of Korea}
\acmISBN{979-8-4007-2259-2/2026/08}
\acmDOI{10.1145/3770855.3817453}

\begin{document}

\title{FUSEP: A Multi-Center Benchmark for Diverse Tasks in Early Pregnancy Fetal Ultrasound Screening}

\author{Bin Pu}
\affiliation{%
  \institution{Hunan University}
  \city{Changsha}
  \state{Hunan}
  \country{China}}
\email{pubin@hnu.edu.cn}

\author{Jiewen Yang}
\affiliation{%
  \institution{The Hong Kong University of Science and Technology}
  \city{HKSAR}
  \country{China}}
  \email{jyangcu@connect.ust.hk}

\author{Liwen Wang}
\affiliation{
 \institution{Anhui University}
 \city{Hefei}
 \state{Anhui}
 \country{China}}
\email{liwenwang@stu.ahu.edu.cn}

\author{
Ying Tan}
\affiliation{%
  \institution{Shenzhen Maternity and Child Healthcare Hospital}
  \city{Shenzhen}
  \state{Guangdong}
  \country{China}}
\email{tanying8013@126.com}

\author{Guannan He}
\affiliation{%
  \institution{Sichuan Provincial Maternity and Child Health Care Hospital}
  \city{Chengdu}
  \state{Sichuan}
  \country{China}}
\email{guannanhe1981@gmail.com}

\author{Xingbo Dong}
\affiliation{%
 \institution{Anhui University}
 \city{Hefei}
 \state{Anhui}
 \country{China}}
\email{xingbo.dong@ahu.edu.cn}

\author{Qika Lin \Letter} \thanks{\Letter Corresponding authors}
\affiliation{%
  \institution{National University of Singapore}
  \city{Singapore}
  \country{Singapore}}
\email{qikalin@foxmail.com}

\author{Jiarong Guo}
\affiliation{%
  \institution{The Hong Kong University of Science and Technology}
  \city{HKSAR}
  \country{China}}
\email{jguoaz@connect.ust.hk}

\author{Lixian Yang}
\affiliation{%
  \institution{Yunnan Maternal and Child Health Hospital}
  \city{Kunming}
  \state{Yunnan}
  \country{China}}
  \email{17787284986@163.com}

\author{Zuozhu Liu}
\affiliation{%
  \institution{Zhejiang University}
  \city{Hangzhou}
  \state{Zhejiang}
  \country{China}}
  \email{zuozhuliu@intl.zju.edu.cn}

\author{Shengli Li}
\affiliation{%
  \institution{Shenzhen Maternity and Child Healthcare Hospital}
  \city{Shenzhen}
  \state{Guangdong}
  \country{China}}
  \email{lsl13530386700@126.com}

\author{Kenli Li \Letter}
\affiliation{%
  \institution{Hunan University}
  \city{Changsha}
  \state{Hunan}
  \country{China}}
  \email{lkl@hnu.edu.cn} 

\renewcommand{\shortauthors}{Bin Pu et al.}

\begin{abstract}
%
A large number of infants with congenital anomalies are born each year globally, especially in areas with underdeveloped medical resources.
Currently, fetal ultrasound screening is the most common modality for early pregnancy anatomy detection. This modality can detect anomalies earlier and provide opportune treatment advice.
However, the lack of an ultrasound dataset on early fetal gestation has slowed down the development of automated assisted diagnosis. 
In this work, we present a benchmark dataset for \emph{\underline{\textbf{F}}}etal \emph{\underline{\textbf{U}}}ltrasound \emph{\underline{\textbf{S}}}creening in \emph{\underline{\textbf{E}}}arly \emph{\underline{\textbf{P}}}regnancy to facilitate intelligent ultrasound examination and assisted diagnosis called \textbf{FUSEP}. 
Our dataset consists of two ultrasound views recommended by the international guideline, i.e., Crown-rump Length (CRL) and Nuchal Translucency (NT) views in three hospitals, totaling 4,017 ultrasound images, with 45,820 box-level expert-level annotations.
Our dataset and baseline present the following three contributions:
1) Our medical experts annotated a total of 14 key anatomical structures in two views using a box-level format;
2) Our data is collected extensively from different sonographers, devices, scanning angles, hospitals, etc;
3) We report the performance of the semi-supervised learning, fully supervised learning, unsupervised domain adaptation (UDA), and source-free UDA in ultrasound images multi-object detection.
To the best of our knowledge, this is the first publicly available dataset and benchmark for fetal early pregnancy ultrasound screening.
We believe that FUSEP and benchmark can contribute to the medical community in the development of multiple tasks such as standard plane recognition, quality control on ultrasound images, automated assisted diagnostics in early fetal pregnancy, medical multi-object detection, domain adaptation for object detection, etc.
\end{abstract}



\maketitle

\section{Introduction}
\label{sec:intro}
%
The birth defect is a significant global problem that affects a large and costly number of newborns each year.
Globally, an estimated 7.9 million infants are born with birth defects \cite{christianson2005march}.
In 2016, the infant death rate in the United States was 5.9 deaths per 1,000 live births, and congenital disabilities were the leading cause of infant deaths, accounting for 20\% of all infant deaths \cite{murphy2018mortality}.
Ultrasonography, as a low-cost, non-invasive, real-time imaging technique, is the primary fetal congenital anomaly detection tool worldwide \cite{pu2021automatic, pu2022mobileunet, zhao2022ultrasound, zhao2024farn}.

In the clinical practice of antenatal ultrasound screening, early pregnancy screening examination plays an important role in enabling early detection of abnormalities for prompt treatment and intervention, which can greatly contribute to fetal health assessment and development.
For example, Michailidis et al. \cite{michailidis2002assessment} found that 93.7\% of the complete anatomy of the fetus could be shown on an early pregnancy ultrasound using two-dimensional ultrasound.
In addition, \cite{jones2006recognizable} reported that more than 80\% of fetal malformations are manifested in early pregnancy before the 12th gestational week.
Various studies \cite{deng2012hierarchical, sciortino2017automatic, ryou2019automated,  sciortino2016wavelet, walker2022using, lin2022much, sonia2015image, nie2017automatic, carneiro2008detection, looney2021fully} highlight the value and importance of the fetal screening examination in the early stage of pregnancy.

\begin{table*}[!t]
    \centering
    \caption{The comparison of our dataset, CardiacUDA~\cite{yang2023graphecho}, CAMUS~\cite{leclerc2019deep}, and EchoNet~\cite{ouyang2020video}.}
    \begin{adjustbox}{width=0.7999\textwidth}
    {\begin{tabular}{c|c|c|c|c}
    \hlineB{3}
    Dataset      &   \textbf{Ours FUSEP (Fetus)}    &  CardiacUDA (Adult)           & CAMUS (Adult)               & EchoNet  (Adult)            \\ \hline
   
    Annotated Images &  4,017  & 4,960& 1,000& 1,755,250 \\
    Ultrasound Views  &   2   & 4                    & 1                    & 1                    \\
    Resolution   &  872-880p   & 720p                & 480p                 & 120p                \\  
    Annotated Regions &   \makecell{NB, NTAPS, MX, MDS, \\
    DP, RBP, LV, H, C,  \\
    AB, G, B, NT, MLS}     & LV, RV, LA, RA                & LV, LA                 & LV                 \\ 
    Number of Centers &   3    & 2 & 1 & 1 \\ 
    \hlineB{3}
    \end{tabular}}
    \end{adjustbox}
    \label{tab:datacompar}
\end{table*}
\begin{figure*}[t!]
    \centering
    \includegraphics[width=0.7999\textwidth]{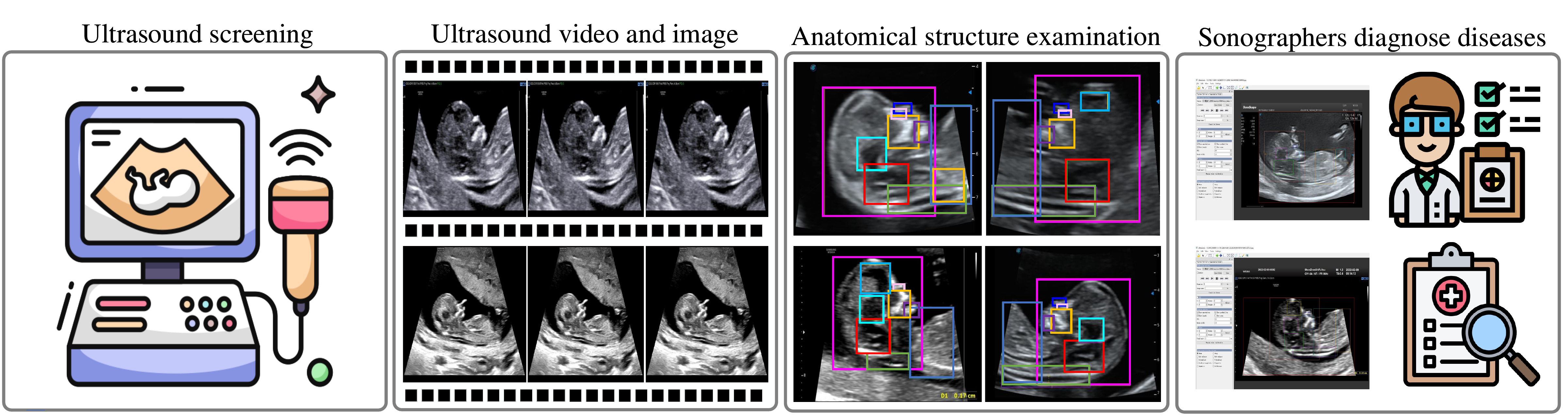}
    \caption{Diagnosis process based on the anatomical structure detection and evaluation.}
    \label{fig:motivation_benchmark}
\end{figure*}

According to the guideline \cite{bilardo2023isuog} of the International Society of Ultrasound in Obstetrics
and Gynecology (ISUOG), early pregnancy screening is mainly focused on the 11-14 gestational weeks of the fetus.
ISUOG pointed out that the Nuchal Translucency view (NT, Figure \ref{fig:CRL_scan}) and the Crown-rump length view (CRL, Figure \ref{fig:CRL_scan}) are particularly important in early pregnancy examinations. These two views not only help accurately measure fetal growth indicators such as nuchal translucency thickness but also diagnose a wide range of serious structural anomalies.
Figure~\ref{fig:motivation_benchmark} illustrates the workflow for examination based on anatomical structural anomalies in early pregnancy. 
Specifically, sonographers first perform ultrasound examinations of the pregnant women. This process produces a series of ultrasound video streams of different views that the sonographer needs to recognize and locate to standard views in early pregnancy for the diagnosis of fetal congenital abnormalities manually.
Secondly, sonographers assess the presence or absence of structures, i.e., anatomical structure examination or growth measurements as shown in Figure \ref{fig:motivation_benchmark}, to determine whether fetuses have abnormalities.
For example, the absence of a nasal bone structure may indicate fetal chromosomal disorders or cardiac malformations \cite{pan2023association}. As another example, when evaluating growth measures, the nuchal translucency structure thickness should be less than 0.25cm in a healthy fetus \cite{sun2021application}; otherwise, an abnormality may occur.

The ultrasound examinations mentioned above rely on the experience level of the sonographer, and internal observations can vary significantly. In addition, ultrasound images are subject to dark shadows, noise, and discontinuities of anatomical structures due to imaging limitations of ultrasound devices, which may lead to an inaccurate diagnosis \cite{drukker2022clinical}.
On the other side, early pregnancy ultrasound examinations are particularly challenging because fetal organs are tiny and difficult to identify, unlike in second and third-trimester examinations.
Those challenges lead to a high rate of underdiagnosis and misdiagnosis \cite{liao2021routine, dawood2022imaging}.
For example, in early pregnancy clinical screening, the detection rate of structural malformations was only 43.1\% (95\% CI 40.6\%-45.5\%) \cite{liao2021routine}.
Therefore, there is an urgent clinical need for robust studies on assisted diagnostics for early pregnancy ultrasounds, as these can efficiently detect and diagnose fetal anatomy during this critical stage of development.

Recently, deep learning has made tremendous progress and gained popularity in medical image recognition \cite{van2022explainable, fiorentino2023review, pu2021automatic, pu2022mobileunet, zhao2022ultrasound, zhao2024farn}, such as classification \cite{pu2021automatic}, segmentation \cite{pu2022mobileunet}, and detection \cite{zhao2022ultrasound, zhao2024farn}.
The use of deep learning techniques to reduce the workload of sonographers and aid diagnosis in ultrasound image analysis has been investigated in \cite{yang2023graphecho, leclerc2019deep}. 
However, research in the aforementioned area remains underexplored or overlooked in the context of early fetal pregnancy ultrasound due to the lack of fetal ultrasound data during early pregnancy.
To this end, we propose two fetal ultrasound views ( CRL and NT) in early pregnancy, along with a benchmark dataset consisting of 4,017 images from three hospitals with a very detailed annotation for anatomical structure.
See Table \ref{tab:datacompar} for a comparison of different ultrasound public datasets.
\textbf{The contributions can be summarised as:}
\\
\noindent
\textbf{1.} We propose a new benchmark dataset for early fetal screening consisting of CRL and NT views across \textbf{three different medical centers}. The ultrasound experts annotated 14 anatomical structures on this dataset, which is, to our knowledge, the most detailed structural annotation for an early pregnancy study. Finally, we give the most recent various detection-based baselines for the anatomical structures. 
\\
\noindent
\textbf{2.} Our dataset can contribute to a range of studies, such as ultrasound clinical tasks (e.g., standard plane recognition \cite{pu2021automatic,zhao2024farn,baumgartner2017sononet,yu2017deep}, quality control \cite{dong2019generic, lin2019multi}, and disease diagnosis \cite{walker2022using}) and computer-based techniques (e.g., medical multi-object detection, unsupervised domain adaptation object detection, and source-free unsupervised domain adaptation object detection in medical image). {Furthermore, FUSEP establishes a rigorous technical testing standard evaluating models across Extreme Scale Variation, Domain Shift, Class Imbalance, Feature Sparsity, and Topological Consistency.} FUSEP can be accessed at: {https://github.com/LiwenWang919/FUSEP}.

\section{Related Work}
%
\subsection{Early Pregnancy Intelligence Screening and Fetal ultrasound}
%
Recently, several studies \cite{deng2012hierarchical, sciortino2017automatic, ryou2019automated, sciortino2016wavelet, walker2022using, lin2022much, sonia2015image, nie2017automatic, carneiro2008detection, looney2021fully} have been conducted to explore intelligent ultrasound screening in early pregnancy, focusing mainly on the NT View (i.e., detection of the nuchal translucency structure) \cite{deng2012hierarchical, sciortino2017automatic}.
For example, Deng et al. \cite{deng2012hierarchical} proposed a hierarchical automated detection model to identify the nuchal translucency structure, head, and body of fetuses, respectively, by spatial constraints and Gaussian pyramids.
Then, Sciortino et al. \cite{sciortino2017automatic} developed an automated method for detecting and measuring nuchal translucency structure thickness in combination with clinical protocols.
Lin et al. \cite{lin2022much} detected 9 anatomical structures based on a RetinaNet detector \cite{lin2017focal} to determine whether the early pregnancy views were standard based on the presence of the structures.
Several learning-based studies \cite{looney2021fully, ryou2019automated, nie2017automatic, yang2018towards} have also investigated the potential of 3D ultrasound for screening in early pregnancy.
Looney et al. \cite{looney2021fully} proposed a novel multiclass (MC) convolutional neural network (CNN) to recognize and segment the placenta, amniotic fluid, and fetus for early pregnancy assessment.
Yang et al. \cite{yang2018towards} designed a fully automated segmentation framework via composite architecture (i.e., 3D CNN and multi-directional recurrent neural network (RNN)) and hierarchical deep supervision mechanism.

There have also been several studies of fetal intelligence ultrasound in the second and third trimesters.
\cite{drukker2021transforming} presents a novel ultrasound acquisition system as well as a coarse-grained automatic annotation deep spatio-temporal network in ultrasound videos for the classification tasks, such as Torax-heart, Head-brain, and Spine. 
\cite{alzubaidi2023large} focuses on the rough classification of sections, such as the brain, abdomen, and thorax, as well as the cavum septi pellucid and the lateral ventricle segmentation of the brain.
FPUS23 \cite{srinivas2023fpus23} is used to identify the correct diagnostic planes for further estimating fetal biometric values in the second trimester.
\cite{chen2024psfhs} focuses on artificial intelligence-based segmentation of pubic symphysis and fetal head in the third trimester, which can accurately assess fetal head descent and predict the most appropriate mode of delivery.
However, most of the previous methods focused only on the NT view, or nuchal translucency structure, ignoring the CRL view and the multiple anatomical structures of early pregnancy, which play an important role in assessing healthy fetal development in early pregnancy.
Secondly, some second and third-trimester fetal ultrasound studies do not involve very detailed identification of multiple anatomical structures.
Thirdly, due to the lack of publicly available data, the most advanced technological research on learning-based screening in early pregnancy has not been fully explored, and previous methods may have been relatively outdated.
Hence, the proposed dataset and benchmark can effectively fill this gap and promote fetal ultrasound early pregnancy learning-based screening, benefiting the entire medical image analysis community.
%
\subsection{Multi-Structure Object Detection in Medical Image Analysis}
%
Multi-structure object detection in medical image analysis is a critical task aimed at identifying and localizing various anatomical structures or pathological findings in medical images~\cite{zhao2022ultrasound,pu2024hfsccd,zhao2024farn, dong2019generic,xu2018less}. This task faces great challenges due to the variable appearance of organ structures, noise, and different imaging principles in different modalities.
Dong~\textit{et al.} proposed a heart structure detection model by integrating the single-shot multi-box detector with aggregated residual visual blocks~\cite{dong2019generic}. Xu~\textit{et al.} presented a framework that integrates global convolutional kernels, coordinate constraints, and a conditional adversarial module to distinguish different structures~\cite{xu2018less}. Additionally, USPD~\cite{zhao2022ultrasound} integrates plane classification and key structure detection for fetal head examinations, offering robust performance against image interference, clinical-grade detection speed, and explanatory decision-making. It enhances anatomical structure recognition through the adaptive integration of hybrid knowledge for improved semantic consistency and reasoning. Although some studies have been done on medical multi-object detection, there is still a lack of publicly available datasets on this research topic.
Besides, related works in fetal ultrasound detection have presented several novel detection methods. FARN~\cite{zhao2024farn} incorporates global context semantic blocks and local topology relationship blocks to model anatomical correlations between different structures in fetal ultrasound images. 

However, as we mentioned, most of these methods are based on in-house datasets and do not release datasets.
%
\subsection{Object Detection in Natural Scenario}
%
In natural scenarios, object detection involves two main tasks: object localization and object classification.
To address these tasks, various deep learning methods have been proposed~\cite{ren2016faster, carion2020endtoend, peize2020sparse, zhu2021deformable, meng2021-CondDETR, liu2022dabdetr, li2022exploring, zhou2022detecting, zong2023detrs, zhang2022dino}. 
These methods can be broadly categorized into two types: two-stage detection and one-stage detection. 
Two-stage detection methods~\cite{ren2015faster, mok2020fast, girshick2015fast, peize2020sparse} provide the advantage of higher accuracy by separating the object detection process into two distinct stages: region proposal generation and object classification. In the first stage, a region proposal network (RPN) generates a set of candidate object bounding boxes. These proposals are then refined and classified in the second stage.
Faster R-CNN~\cite{ren2015faster} is one of the most well-known two-stage detectors, which integrates the RPN for end-to-end training. This method achieves high accuracy by focusing on the most promising regions of an image.
On the other hand, one-stage detection improves the detection process by combining object localization and classification into a single step, which leads to faster inference speeds. YOLO~\cite{redmon2016you} is designed as the pioneering framework for one-stage detection, offering real-time processing capabilities by directly predicting bounding boxes and class probabilities from the entire image in a single evaluation. 
DETR (Detection Transformer)~\cite{carion2020endtoend} introduces a transformer-based approach that simplifies the detection pipeline by eliminating the need for region proposals, with Deformable DETR~\cite{zhu2021deformable} enhancing this model through deformable attention modules for better performance and faster training. Conditional DETR~\cite{meng2021-CondDETR} and DAB-DETR~\cite{liu2022dabdetr} refine transformer-based detection with conditional cross-attention and dynamic anchor boxes, while vision transformer-based detectors like ViTDet~\cite{li2022exploring} and GLIP~\cite{li2022grounded} utilize large-scale pre-training for enhanced generalization and accuracy.
Compared with two-stage detection, one-stage detection provides a lower inference time in real-world scenarios, especially in early pregnancy anatomical detection cases.

\section{Cohort Definition, Dataset Composition \& Dataset Significance}
%

\begin{figure*}[t!]
    \centering
    \begin{minipage}[b]{0.58\textwidth} 
        \centering
        \includegraphics[width=\linewidth]{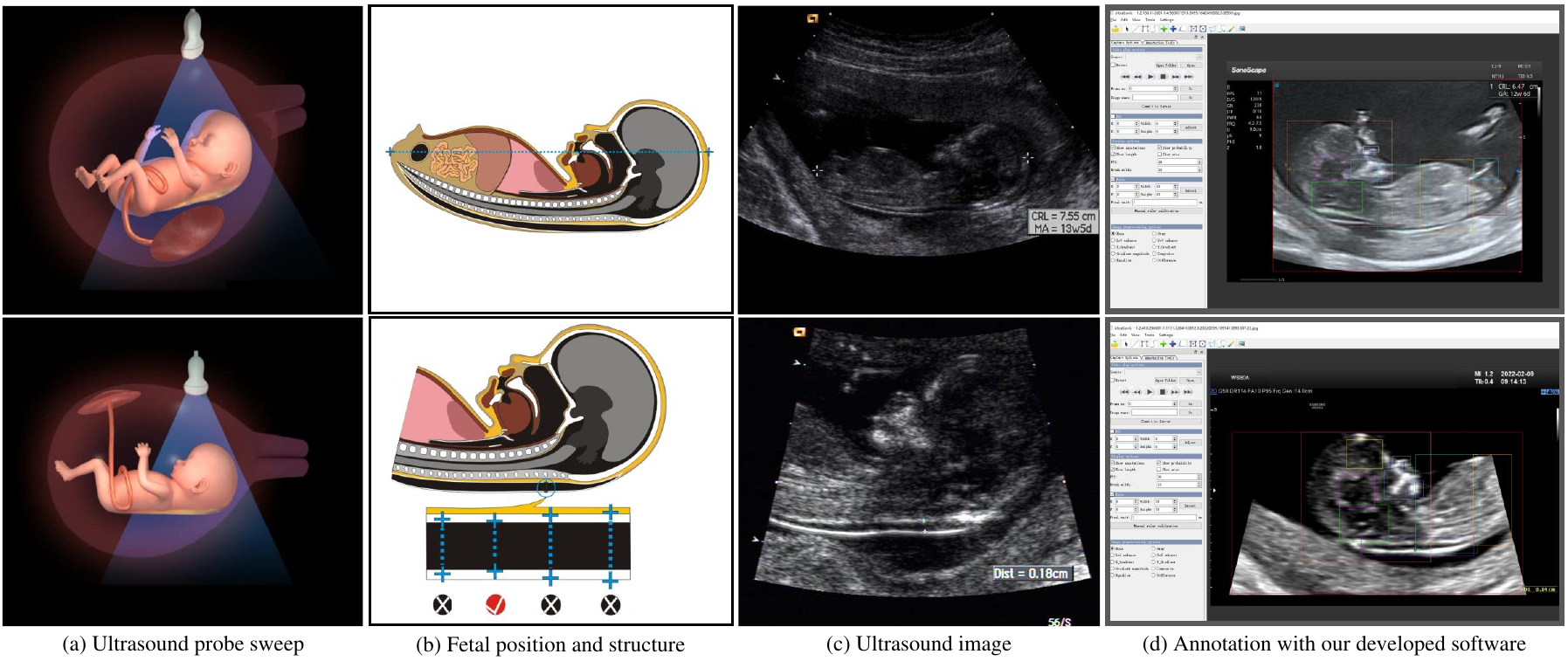}
        \caption{CRL and NT views scanning process and annotation procedure.}
        \label{fig:CRL_scan}
    \end{minipage}
    \hfill 
    \begin{minipage}[b]{0.38\textwidth} 
        \centering
        \begin{adjustbox}{width=\linewidth} 
            \begin{tabular}{cc|cc}
\hline
\multicolumn{2}{c|}{CRL View}                & \multicolumn{2}{c}{NT View}                    \\ \hline
\multicolumn{1}{c|}{\textbf{Structure}}          & \textbf{Abb} & \multicolumn{1}{c|}{\textbf{Structure}}          & \textbf{Abb} \\ \hline
\multicolumn{1}{c|}{Maxilla}           & MX  & \multicolumn{1}{c|}{Maxilla}             & MX  \\
\multicolumn{1}{c|}{Mandible with a Dot Shape}   & MDS          & \multicolumn{1}{c|}{Mandible with a Dot Shape}   & MDS          \\
\multicolumn{1}{c|}{Mandible with Long Strip}    & MLS          & \multicolumn{1}{c|}{Mandible with Long Strip}    & MLS          \\
\multicolumn{1}{c|}{Lateral Ventricle} & LV  & \multicolumn{1}{c|}{Lateral Ventricle}   & LV  \\
\multicolumn{1}{c|}{Head}              & H   & \multicolumn{1}{c|}{Head}                & H   \\
\multicolumn{1}{c|}{Genitals}          & G   & \multicolumn{1}{c|}{Chest}               & C   \\
\multicolumn{1}{c|}{Chest}             & C   & \multicolumn{1}{c|}{ABdomen}             & AB  \\
\multicolumn{1}{c|}{ABdomen}           & AB  & \multicolumn{1}{c|}{RhomBencePhalon}     & RBP \\
\multicolumn{1}{c|}{Breech}            & B   & \multicolumn{1}{c|}{Dience Phalon}        & DP  \\
\multicolumn{1}{c|}{RhomBencePhalon}   & RBP & \multicolumn{1}{c|}{Nuchal Translucency} & NT  \\
\multicolumn{1}{c|}{Dience Phalon}                & DP           & \multicolumn{1}{c|}{Nasal Tip And Prenasal Skin} & NTAPS        \\
\multicolumn{1}{c|}{Nasal Tip And Prenasal Skin} & NTAPS        & \multicolumn{1}{c|}{Nasal Bone}                  & NB           \\
\multicolumn{1}{c|}{Nasal Bone}        & NB  & \multicolumn{1}{c|}{}                    &     \\ \hline
\end{tabular}
        \end{adjustbox}
        \captionof{table}{\MakeUppercase Key structure and corresponding abbreviation (Abb).}
        \label{tab:abb}
    \end{minipage}
\end{figure*}

\subsection{Dataset Collection.}
\label{sec:collect}

\textbf{Cohort Definition and Fetus Ultrasound Scanning Standards.} 
\label{sec:ultrasound_scan_standards}
This is a retrospective study, and the main target population for this dataset is pregnant women undergoing early pregnancy screening of fetal structures between 11 and 14 gestational weeks in a hospital ultrasound department.
All collected cases in our dataset have no obvious abnormalities found in subsequent ultrasound examinations and postnatal follow-ups.
According to the guideline~\cite{bilardo2023isuog}  for early pregnancy fetal examination, the routine screening will retain approximately 20 relevant views. Those views are crucial for measuring fetal growth parameters and screening for congenital diseases. Among those views, the fetal Nuchal Translucency view (NT) and Crown–rump length View (CRL) are the most important for diagnosing various severe structural anomalies. {For example, the measurement of NT thickness and the detection of the nasal bone in these views are crucial early markers for chromosomal abnormalities such as Down syndrome, while CRL is a primary metric for monitoring early fetal growth and development. FUSEP is designed as a foundational benchmark that prioritizes these two highest-yield views, and we plan to expand to additional views in future iterations.}
For the scanning of NT and CRL view, we follow the international early pregnancy ultrasound guideline~\cite{bilardo2023isuog} and have the following standards for NT and CRL views:
\\
\noindent
\textbf{Scanning of NT}: As shown in the \emph{top row} of Figure~\ref{fig:CRL_scan}, we enlarge the image to display only the fetal head and upper chest. The scanned fetus is in a natural position, with amniotic fluid visible between the chin and chest wall. The scanned image is in midsagittal shape, with a clear view of the nasal tip. The fetal head and maxilla form a rectangular shape. The diencephalon appears transparent, and the transparent layer behind the neck is clear in the image.
\\
\noindent
\textbf{Scanning of CRL}: As shown in the \emph {bottom row} of Figure~\ref{fig:CRL_scan}, we enlarge the image to display the whole body of the fetus in a natural position. The scanned image is in midsagittal shape, and the Crown-rump line should be parallel to the horizon to the best possible.
\begin{figure*}[t!]
    \centering
    \includegraphics[width=0.8499\textwidth]{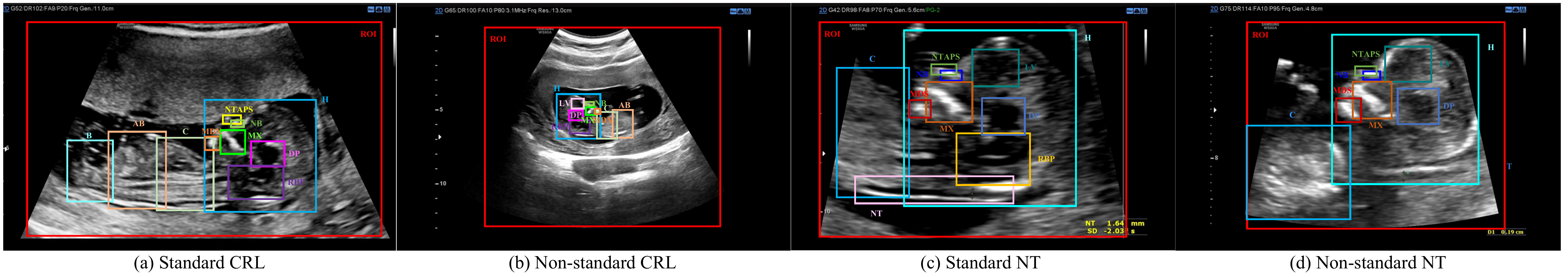}
    \caption{Examples of our datasets FUSEP for the standard and non-standard CRL and NT views.}
    \label{fig:dataset_examples}
\end{figure*}

Finally, the collected image contains the key structures presented in Table~\ref{tab:abb} of the fetus to the fullest extent possible. It is usually challenging for sonographers to obtain images under the ideal conditions described above. Specifically, some common situations may affect the quality of scanned images, such as changes in posture or the development of the fetus not yet reaching the standard in the early pregnancy stage, which may result in some organs not being detected or scanned. In order to effectively monitor and track fetal development at later stages, the hospital archives all ultrasound images for subsequent examination and patient follow-up.

To construct a comprehensive and diverse fetal ultrasound dataset that supports the development and evaluation of deep learning models, data were collected from \textbf{three distinct medical centers}. This multi-center dataset enhances the generalizability of the dataset by capturing variations in scanning protocols, patient demographics, and sonographer practices. The study included cases from three medical centers (\textbf{Hospital 1}, \textbf{Hospital 2}, and \textbf{Hospital 3}), with 766, 759, and 989 cases, and 1,532, 1,496, and 989 ultrasound images, respectively. 
The number of anatomical structures labeled at the box-level for Hospitals 1, 2, and 3 are, respectively, 13,485, 14,281, and 18,054.
However, to reflect real-world clinical settings and enrich the dataset's utility across multiple tasks, such as standard view recognition, image quality assessment, and anatomical structure detection, not all acquired views strictly adhere to standardized criteria. 

Furthermore, to enable domain adaptation research, particularly cross-device adaptation, images were acquired using different ultrasound devices: SAMSUNG, Sonoscape, and GE. {Our analysis shows that domain shifts are primarily driven by the use of these fundamentally different ultrasound devices across centers, as well as varying class area ratios, which directly accounts for the varying difficulty in cross-center tasks.} This inter-device variability introduces additional domain shifts that are critical for evaluating and improving the robustness of machine learning models in clinical settings. All data collection and experimental procedures were conducted in accordance with the guidelines of the local ethics committee and received formal approval (LLYJ2024-370-121). 

\textbf{Fetus Ultrasound Image Collection Standards and Pipeline for Anatomy Detection.}
%
During practical ultrasound image collection, sonographers first ensure that the fetus is in a natural position so that most of the structures are visually available. Also, the scanned image should be in mid-sagittal shape and clearly show the structures (As examples shown in Figure~\ref{fig:dataset_examples}). The ultrasound device performs a real-time video capture process, and the ultrasound probe may also have slight movement due to the actions of the sonographers during scanning, which may cause motion blur and reduce the quality of the captured images. To address this problem, the sonographers will select an image from the captured video with the best quality. The next step will be to conduct further annotations of selected images.

%
\subsection{Dataset Annotation.}
%
As shown in Figure \ref{fig:CRL_scan}, two sonographers with more than seven years of clinical experience in fetal ultrasound respond to annotating the dataset. To ensure that the annotation process is efficient and high-quality, we also develop software to assist in ultrasound image annotation (see Figure~\ref{fig:CRL_scan}(d) for the example of our software). 
The annotation includes,
view category (e.g., NT or CRL), the key structures displayed in scanned views, and the standardization (standard/non-standard) of collected images. 
If it is a standard view, it implies that the image can be used directly for diagnosis as diagnostic evidence, and vice versa if it is non-standard.
For NT, key structures include the nasal tip and anterior nasal skin, nasal bone, maxilla, mandible, diencephalon, rhombic brain, and posterior cervical translucency. For CRL, in addition to the above structures, the fetal genitalia is also labeled. 
All samples included in the study were annotated consistently by two experienced ultrasonographers following established international ultrasound guidelines~\cite{salomon2019isuog}. Any inconsistent annotations were identified and excluded during the preprocessing stage to ensure high-quality labeling. The inter-annotator agreement was guided by standardized criteria from these guidelines, ensuring reliable and reproducible annotation across all samples. 

In our dataset, each image is accompanied by metadata stored in multiple formats: JPG for image data, Excel for fetal age information, and JSON for detailed annotation details. Specifically, each JSON file contains four types of annotations: (1) the image category (e.g., NT or CRL), (2) box-level coordinates of multiple anatomical structures, with four vertices per structure, and (3) ROI (region of interest) coordinates, although ROI is not used in this study. This structured annotation approach ensures clarity, consistency, and usability for downstream analysis.

\subsection{Dataset Significance in Medical Community.}
Our dataset can contribute to at least six following research topics. \textbf{Standard view recognition, quality control, and potential disease diagnosis based on missing anatomical structures} are three clinical tasks that can be explored and addressed on the basis of object detection approaches, which include supervised and semi-supervised methods. Furthermore, we address the domain adaptation problem through two distinct experimental frameworks: \textbf{unsupervised domain adaptation (UDA) for detection tasks}, and \textbf{source-free domain adaptation (SFDA) where only source model parameters are accessible.} The following description illustrates the details of each task.

\textit{\textbf{A. Standard view recognition}}~\cite{pu2021automatic, zhao2024farn}. Various studies \cite{yu2016fetal, pu2021automatic, zhao2024farn} have highlighted the important role of standard view recognition in obstetric ultrasound. Our dataset contributes significantly to the advancement of intelligent ultrasound systems for standardized image acquisition. This application enhances the quality of prenatal sonography services by enabling sonographers to focus more on detailed fetal anatomical assessments and early detection of potential abnormalities. It not only improves diagnostic accuracy but also elevates the overall quality of ultrasound examinations. Furthermore, identifying standardized imaging planes provides a necessary foundation for subsequent automated measurements and data analysis, such as nuchal translucency (NT) and crown-rump length (CRL) measurements. Ensuring the precision and reproducibility of these quantitative parameters offers robust support for clinical decision-making.
\\
\textit{\textbf{B. Quality control on ultrasound images}}~\cite{dong2019generic, lin2019multi, zhang2021automatic}.
Previously, experienced sonographers were required to evaluate ultrasound images acquired by junior sonographers to assess their competency, which increased workload and introduced inter-observer variability. Hence, there is a pressing need for publicly available datasets to advance automated image quality assessment. Our dataset serves as a valuable resource for developing intelligent ultrasound quality control systems. These systems can objectively assess the proficiency of medical personnel in acquiring standardized views, thereby streamlining training and ensuring consistent image quality across operators and institutions.
\\
\textit{\textbf{C. {Preliminary Extension for Potential disease diagnosis} based on missing anatomical structures}}~\cite{walker2022using}.
{Our dataset serves primarily as an anatomical detection and screening dataset, but supports preliminary exploration of diagnostic tools} for detecting structural anomalies during early pregnancy. For instance, the absence of the fetal nasal bone, detectable via ultrasound in early gestation, is strongly associated with chromosomal abnormalities such as Down syndrome \cite{carvalho2023isuog}. Early identification of such markers can provide expectant parents timely access to further diagnostic testing, aiding in confirming or excluding genetic disorders \cite{karim2024detection}. This early intervention not only helps reduce the incidence of birth defects but also empowers families to make informed reproductive decisions.
\\
\textit{\textbf{D. Medical multi-object detection}}~\cite{baumgartner2017sononet, nguyen2021circle}.
Multi-object detection in medical imaging forms the basis for early diagnosis, improved diagnostic efficiency, reduced missed or incorrect diagnoses, comprehensive patient evaluation, and decision support for personalized treatment strategies. However, no publicly accessible dataset currently supports the detection of multiple anatomical structures in early pregnancy ultrasound. Our dataset includes annotations from two standard views and covers 14 key anatomical structures, making it suitable for advancing methodologies such as fully supervised and semi-supervised object detection techniques tailored to early pregnancy ultrasound.
\\
\textit{\textbf{E. Semi-supervised learning in multi-object detection.}}
In medical imaging, especially early pregnancy ultrasound, large annotated datasets are scarce. Semi-supervised learning leverages both labelled and unlabeled data, improving model generalization when annotations are limited. Data augmentation further complements SSL, which introduces variability through transformations, enhancing robustness. Our proposed FUSEP dataset, with rich annotations of 14 structures across two views, provides an ideal basis for semi-supervised learning and augmentation, enabling accurate fetal parameter detection and improved prenatal diagnosis.
\\
\textit{\textbf{F. Domain adaptation for object detection in medical image analysis.}}~\cite{liu2023decoupled, liu2022source}.
As noted in \cite{liu2023decoupled, liu2022source}, domain shifts commonly occur between data collected from different hospitals or imaging devices. These shifts often cause performance degradation when deep learning models trained on source-domain data are applied to target-domain data. Domain adaptation has proven effective in addressing this challenge \cite{oza2023unsupervised}. Our dataset was acquired using different distinct ultrasound platforms, SAMSUNG and Sonoscape, which naturally introduce domain gaps due to differences in imaging characteristics. Thus, this dataset is well-designed for evaluating and advancing domain adaptation techniques in medical image analysis, particularly in object detection tasks.

{Furthermore, this benchmark drives the community to explore new research questions: (1) How to maintain Topological Consistency among anatomical structures when detecting highly sparse features (e.g., Feature Sparsity in early fetal development); (2) How to design source-free unsupervised domain adaptation (SFDA) methods that are resilient to cross-device ultrasound artifacts; and (3) How to effectively utilize semi-supervised learning when expert annotations for specific rare structures (e.g., absent nasal bones) are extremely scarce.}

\begin{table*}[t!]
\centering
\caption{Quantitative fully supervised results for CRL and NT. The mAP is reported in each cell in the order Hospital 1/ 2/ 3.}
\setlength{\tabcolsep}{2pt}
\begin{adjustbox}{width=0.9999\textwidth}
    \begin{tabular}{ccccccccccccccc}
\hline
\multicolumn{1}{c|}{Method} &
  \multicolumn{1}{c|}{mAP  (\%)} &
  \multicolumn{1}{c|}{MX} &
  \multicolumn{1}{c|}{MDS} &
  \multicolumn{1}{c|}{MLS} &
  \multicolumn{1}{c|}{LV} &
  \multicolumn{1}{c|}{H} &
  \multicolumn{1}{c|}{G} &
  \multicolumn{1}{c|}{C} &
  \multicolumn{1}{c|}{AB} &
  \multicolumn{1}{c|}{B} &
  \multicolumn{1}{c|}{RBP} &
  \multicolumn{1}{c|}{DP} &
  \multicolumn{1}{c|}{NTAPS} &
  NB \\ \hline
\multicolumn{15}{c}{CRL} \\ \hline
\multicolumn{1}{c|}{Faster-RCNN \cite{ren2016faster}} &
  \multicolumn{1}{c|}{86.3/92.6/82.7} &
  \multicolumn{1}{c|}{93.3/96.4/90.5} &
  \multicolumn{1}{c|}{92.7/93.1/75.9} &
  \multicolumn{1}{c|}{75.2/74.0/64.0} &
  \multicolumn{1}{c|}{78.4/82.3/71.5} &
  \multicolumn{1}{c|}{98.2/98.6/92.0} &
  \multicolumn{1}{c|}{79.7/92.5/71.9} &
  \multicolumn{1}{c|}{95.5/97.2/93.6} &
  \multicolumn{1}{c|}{97.4/99.4/92.3} &
  \multicolumn{1}{c|}{98.2/98.2/91.9} &
  \multicolumn{1}{c|}{93.7/96.6/89.3} &
  \multicolumn{1}{c|}{97.7/97.4/92.5} &
  \multicolumn{1}{c|}{79.9/93.5/73.8} &
  42.6/84.6/76.0 \\
\multicolumn{1}{c|}{DETR \cite{carion2020endtoend}} &
  \multicolumn{1}{c|}{90.0/93.5/83.5} &
  \multicolumn{1}{c|}{91.7/97.9/91.1} &
  \multicolumn{1}{c|}{90.7/95.6/74.0} &
  \multicolumn{1}{c|}{74.3/74.5/65.9} &
  \multicolumn{1}{c|}{87.7/89.4/74.5} &
  \multicolumn{1}{c|}{97.4/98.7/92.4} &
  \multicolumn{1}{c|}{87.4/91.2/70.4} &
  \multicolumn{1}{c|}{95.9/99.8/94.0} &
  \multicolumn{1}{c|}{97.3/99.4/93.2} &
  \multicolumn{1}{c|}{97.9/98.4/94.0} &
  \multicolumn{1}{c|}{94.8/98.2/89.5} &
  \multicolumn{1}{c|}{98.3/98.9/92.2} &
  \multicolumn{1}{c|}{82.2/92.2/77.5} &
  74.3/80.9/75.6 \\
\multicolumn{1}{c|}{YOLOX \cite{yolox2021}} &
  \multicolumn{1}{c|}{92.3/94.0/84.1} &
  \multicolumn{1}{c|}{93.9/99.3/91.8} &
  \multicolumn{1}{c|}{92.9/91.4/76.3} &
  \multicolumn{1}{c|}{77.6/77.9/67.3} &
  \multicolumn{1}{c|}{92.0/86.9/73.3} &
  \multicolumn{1}{c|}{98.9/98.9/93.0} &
  \multicolumn{1}{c|}{89.4/92.8/73.6} &
  \multicolumn{1}{c|}{97.3/99.8/95.2} &
  \multicolumn{1}{c|}{98.3/98.8/94.0} &
  \multicolumn{1}{c|}{98.8/98.0/92.3} &
  \multicolumn{1}{c|}{96.3/98.2/91.5} &
  \multicolumn{1}{c|}{99.0/97.5/92.5} &
  \multicolumn{1}{c|}{87.1/95.7/78.3} &
  78.5/87.4/78.2 \\
\multicolumn{1}{c|}{Deformable DETR \cite{zhu2021deformable}} &
  \multicolumn{1}{c|}{92.7/95.3/84.5} &
  \multicolumn{1}{c|}{94.1/99.3/91.2} &
  \multicolumn{1}{c|}{93.9/97.2/76.3} &
  \multicolumn{1}{c|}{81.6/80.8/66.6} &
  \multicolumn{1}{c|}{91.8/89.9/74.1} &
  \multicolumn{1}{c|}{98.8/98.9/92.3} &
  \multicolumn{1}{c|}{91.1/95.4/73.6} &
  \multicolumn{1}{c|}{97.0/99.2/95.7} &
  \multicolumn{1}{c|}{98.5/98.9/93.3} &
  \multicolumn{1}{c|}{98.7/98.7/95.0} &
  \multicolumn{1}{c|}{96.7/99.0/92.8} &
  \multicolumn{1}{c|}{99.2/97.3/92.9} &
  \multicolumn{1}{c|}{85.9/94.8/78.1} &
  77.0/89.9/75.3 \\
\multicolumn{1}{c|}{ViTDet \cite{li2022exploring}} &
  \multicolumn{1}{c|}{92.9/95.6/84.8} &
  \multicolumn{1}{c|}{93.9/99.7/92.7} &
  \multicolumn{1}{c|}{94.1/96.4/77.3} &
  \multicolumn{1}{c|}{78.4/83.0/67.9} &
  \multicolumn{1}{c|}{93.0/90.4/77.8} &
  \multicolumn{1}{c|}{98.9/99.0/92.9} &
  \multicolumn{1}{c|}{89.1/94.8/71.8} &
  \multicolumn{1}{c|}{97.7/99.9/95.5} &
  \multicolumn{1}{c|}{98.0/99.3/93.8} &
  \multicolumn{1}{c|}{99.8/98.9/96.4} &
  \multicolumn{1}{c|}{97.1/99.0/92.7} &
  \multicolumn{1}{c|}{99.2/98.2/93.4} &
  \multicolumn{1}{c|}{88.0/94.1/76.4} &
  81.0/90.0/72.6 \\
\multicolumn{1}{c|}{CO-DETR \cite{zong2023detrs}} &
  \multicolumn{1}{c|}{92.8/95.7/85.3} &
  \multicolumn{1}{c|}{92.7/98.9/93.1} &
  \multicolumn{1}{c|}{94.0/96.7/75.5} &
  \multicolumn{1}{c|}{76.3/80.6/65.1} &
  \multicolumn{1}{c|}{93.0/92.8/78.4} &
  \multicolumn{1}{c|}{98.9/99.0/93.6} &
  \multicolumn{1}{c|}{90.4/94.5/74.2} &
  \multicolumn{1}{c|}{97.9/100/95.3} &
  \multicolumn{1}{c|}{98.6/99.0/95.2} &
  \multicolumn{1}{c|}{99.9/98.8/95.0} &
  \multicolumn{1}{c|}{96.8/99.1/93.2} &
  \multicolumn{1}{c|}{99.3/98.3/93.9} &
  \multicolumn{1}{c|}{85.6/95.8/78.2} &
  83.2/90.3/76.4 \\
\multicolumn{1}{c|}{DINO \cite{zhang2022dino}} &
  \multicolumn{1}{c|}{93.6/95.6/85.1} &
  \multicolumn{1}{c|}{93.4/98.3/91.7} &
  \multicolumn{1}{c|}{93.2/96.5/77.4} &
  \multicolumn{1}{c|}{75.0/83.0/68.1} &
  \multicolumn{1}{c|}{92.0/91.4/78.3} &
  \multicolumn{1}{c|}{98.7/99.0/93.7} &
  \multicolumn{1}{c|}{92.7/94.7/73.3} &
  \multicolumn{1}{c|}{98.2/100/96.7} &
  \multicolumn{1}{c|}{98.3/99.0/93.6} &
  \multicolumn{1}{c|}{99.9/98.6/95.7} &
  \multicolumn{1}{c|}{96.8/99.2/92.2} &
  \multicolumn{1}{c|}{99.4/98.2/93.4} &
  \multicolumn{1}{c|}{90.0/95.9/78.4} &
  89.3/89.4/75.3 \\
\multicolumn{1}{c|}{DDQ \cite{Zhang_2023_CVPR}} &
  \multicolumn{1}{c|}{93.3/96.1/85.4} &
  \multicolumn{1}{c|}{93.4/98.4/91.8} &
  \multicolumn{1}{c|}{93.7/97.3/78.9} &
  \multicolumn{1}{c|}{76.9/86.2/67.4} &
  \multicolumn{1}{c|}{90.4/92.4/79.0} &
  \multicolumn{1}{c|}{98.4/99.0/93.5} &
  \multicolumn{1}{c|}{92.0/95.0/73.5} &
  \multicolumn{1}{c|}{97.8/99.9/96.2} &
  \multicolumn{1}{c|}{98.7/99.0/94.2} &
  \multicolumn{1}{c|}{99.9/98.8/96.0} &
  \multicolumn{1}{c|}{97.1/99.1/93.1} &
  \multicolumn{1}{c|}{99.4/99.3/93.6} &
  \multicolumn{1}{c|}{89.1/95.3/78.5} &
  85.8/90.3/73.0 \\
\multicolumn{1}{c|}{Relation-DETR \cite{hou2024relation}} &
  \multicolumn{1}{c|}{93.7/96.3/85.6} &
  \multicolumn{1}{c|}{94.1/98.7/93.1} &
  \multicolumn{1}{c|}{94.8/96.7/79.7} &
  \multicolumn{1}{c|}{77.4/86.5/68.4} &
  \multicolumn{1}{c|}{91.5/93.3/78.3} &
  \multicolumn{1}{c|}{98.9/99.0/93.3} &
  \multicolumn{1}{c|}{90.3/94.4/73.8} &
  \multicolumn{1}{c|}{98.2/100/96.4} &
  \multicolumn{1}{c|}{99.1/99.5/94.1} &
  \multicolumn{1}{c|}{99.9/98.8/94.8} &
  \multicolumn{1}{c|}{97.0/99.1/92.4} &
  \multicolumn{1}{c|}{99.3/99.1/94.1} &
  \multicolumn{1}{c|}{90.4/96.8/80.4} &
  86.9/90.3/73.1 \\ \hline
\multicolumn{1}{c|}{Method} &
  \multicolumn{1}{c|}{mAP  (\%)} &
  \multicolumn{1}{c|}{MX} &
  \multicolumn{1}{c|}{MDS} &
  \multicolumn{1}{c|}{MLS} &
  \multicolumn{1}{c|}{LV} &
  \multicolumn{1}{c|}{H} &
  \multicolumn{1}{c|}{C} &
  \multicolumn{1}{c|}{AB} &
  \multicolumn{1}{c|}{RBP} &
  \multicolumn{1}{c|}{DP} &
  \multicolumn{1}{c|}{NT} &
  \multicolumn{1}{c|}{NTAPS} &
  \multicolumn{1}{c|}{NB} &
   \\ \hline
\multicolumn{15}{c}{NT} \\ \hline
\multicolumn{1}{c|}{Faster-RCNN \cite{ren2016faster}} &
  \multicolumn{1}{c|}{88.5/90.2/85.1} &
  \multicolumn{1}{c|}{98.8/97.7/91.9} &
  \multicolumn{1}{c|}{91.1/95.7/83.9} &
  \multicolumn{1}{c|}{60.7/64.9/71.0} &
  \multicolumn{1}{c|}{86.1/89.3/85.4} &
  \multicolumn{1}{c|}{98.1/99.7/95.6} &
  \multicolumn{1}{c|}{94.9/99.0/91.1} &
  \multicolumn{1}{c|}{71.1/63.0/73.9} &
  \multicolumn{1}{c|}{98.3/99.7/94.0} &
  \multicolumn{1}{c|}{97.9/99.8/96.8} &
  \multicolumn{1}{c|}{80.1/94.9/72.7} &
  \multicolumn{1}{c|}{94.8/96.1/85.1} &
  \multicolumn{1}{c|}{89.9/92.8/78.4} & -
   \\
\multicolumn{1}{c|}{DETR \cite{carion2020endtoend}} &
  \multicolumn{1}{c|}{90.7/92.4/88.7} &
  \multicolumn{1}{c|}{99.3/97.3/91.9} &
  \multicolumn{1}{c|}{93.7/95.9/86.8} &
  \multicolumn{1}{c|}{71.8/71.3/73.2} &
  \multicolumn{1}{c|}{92.8/91.1/86.7} &
  \multicolumn{1}{c|}{97.8/99.9/96.7} &
  \multicolumn{1}{c|}{97.8/99.6/91.3} &
  \multicolumn{1}{c|}{80.2/75.5/75.5} &
  \multicolumn{1}{c|}{98.1/99.7/95.9} &
  \multicolumn{1}{c|}{99.1/100/96.9} &
  \multicolumn{1}{c|}{79.8/89.2/77.3} &
  \multicolumn{1}{c|}{91.9/96.4/89.5} &
  \multicolumn{1}{c|}{86.5/93.2/75.7} & -
   \\
\multicolumn{1}{c|}{YOLOX \cite{yolox2021}} &
  \multicolumn{1}{c|}{92.0/95.0/88.5} &
  \multicolumn{1}{c|}{99.3/98.0/93.1} &
  \multicolumn{1}{c|}{93.7/97.0/84.8} &
  \multicolumn{1}{c|}{73.3/77.1/72.7} &
  \multicolumn{1}{c|}{93.5/95.4/85.9} &
  \multicolumn{1}{c|}{97.2/100/97.1} &
  \multicolumn{1}{c|}{98.4/99.9/92.5} &
  \multicolumn{1}{c|}{85.7/89.8/76.8} &
  \multicolumn{1}{c|}{99.1/99.9/96.0} &
  \multicolumn{1}{c|}{99.2/100/96.3} &
  \multicolumn{1}{c|}{77.1/91.5/75.5} &
  \multicolumn{1}{c|}{95.9/97.1/85.0} &
  \multicolumn{1}{c|}{92.0/94.3/78.2} & -
   \\
\multicolumn{1}{c|}{Deformable DETR \cite{zhu2021deformable}} &
  \multicolumn{1}{c|}{90.9/94.8/86.2} &
  \multicolumn{1}{c|}{98.4/98.0/92.3} &
  \multicolumn{1}{c|}{94.0/97.6/86.2} &
  \multicolumn{1}{c|}{72.1/78.8/70.2} &
  \multicolumn{1}{c|}{92.2/95.1/86.1} &
  \multicolumn{1}{c|}{98.0/100/96.7} &
  \multicolumn{1}{c|}{97.9/99.9/92.1} &
  \multicolumn{1}{c|}{85.1/86.5/78.5} &
  \multicolumn{1}{c|}{99.3/99.9/96.9} &
  \multicolumn{1}{c|}{99.5/100/97.1} &
  \multicolumn{1}{c|}{81.8/89.6/71.7} &
  \multicolumn{1}{c|}{96.7/97.6/89.6} &
  \multicolumn{1}{c|}{91.6/94.9/76.2} & -
   \\
\multicolumn{1}{c|}{ViTDet \cite{li2022exploring}} &
  \multicolumn{1}{c|}{92.6/95.0/87.1} &
  \multicolumn{1}{c|}{99.5/97.9/93.3} &
  \multicolumn{1}{c|}{94.9/97.3/87.6} &
  \multicolumn{1}{c|}{75.7/74.4/71.0} &
  \multicolumn{1}{c|}{93.3/96.5/87.7} &
  \multicolumn{1}{c|}{98.0/100/96.3} &
  \multicolumn{1}{c|}{97.9/100/91.9} &
  \multicolumn{1}{c|}{87.2/88.4/76.9} &
  \multicolumn{1}{c|}{99.5/99.9/96.5} &
  \multicolumn{1}{c|}{99.5/100/97.3} &
  \multicolumn{1}{c|}{82.8/92.2/71.5} &
  \multicolumn{1}{c|}{96.3/97.9/89.5} &
  \multicolumn{1}{c|}{91.7/94.9/79.6} & -
   \\
\multicolumn{1}{c|}{CO-DETR \cite{zong2023detrs}} &
  \multicolumn{1}{c|}{91.9/95.1/87.5} &
  \multicolumn{1}{c|}{99.5/97.3/93.0} &
  \multicolumn{1}{c|}{94.6/98.4/87.5} &
  \multicolumn{1}{c|}{71.7/78.2/71.1} &
  \multicolumn{1}{c|}{92.6/94.5/86.5} &
  \multicolumn{1}{c|}{98.0/100/97.2} &
  \multicolumn{1}{c|}{98.4/100/93.0} &
  \multicolumn{1}{c|}{86.2/89.2/77.8} &
  \multicolumn{1}{c|}{99.0/99.9/97.0} &
  \multicolumn{1}{c|}{98.8/100/97.2} &
  \multicolumn{1}{c|}{84.6/91.3/73.0} &
  \multicolumn{1}{c|}{95.3/97.7/88.0} &
  \multicolumn{1}{c|}{92.3/95.3/79.2} &-
   \\
\multicolumn{1}{c|}{DINO \cite{zhang2022dino}} &
  \multicolumn{1}{c|}{91.9/95.2/88.7} &
  \multicolumn{1}{c|}{99.6/97.6/94.0} &
  \multicolumn{1}{c|}{94.5/97.8/84.4} &
  \multicolumn{1}{c|}{70.5/75.9/71.7} &
  \multicolumn{1}{c|}{92.3/96.5/88.9} &
  \multicolumn{1}{c|}{97.0/100/96.7} &
  \multicolumn{1}{c|}{99.0/99.9/93.0} &
  \multicolumn{1}{c|}{87.4/89.4/78.7} &
  \multicolumn{1}{c|}{99.4/99.9/97.3} &
  \multicolumn{1}{c|}{99.0/100/97.2} &
  \multicolumn{1}{c|}{86.1/92.1/75.3} &
  \multicolumn{1}{c|}{95.6/98.3/87.9} &
  \multicolumn{1}{c|}{92.5/95.4/78.7} & -
   \\
\multicolumn{1}{c|}{DDQ \cite{Zhang_2023_CVPR}} &
  \multicolumn{1}{c|}{92.0/95.4/87.4} &
  \multicolumn{1}{c|}{99.6/97.7/93.7} &
  \multicolumn{1}{c|}{94.3/97.6/87.9} &
  \multicolumn{1}{c|}{72.6/75.8/72.0} &
  \multicolumn{1}{c|}{92.9/96.8/88.0} &
  \multicolumn{1}{c|}{98.0/100/96.0} &
  \multicolumn{1}{c|}{98.9/100/93.3} &
  \multicolumn{1}{c|}{87.9/90.8/76.3} &
  \multicolumn{1}{c|}{99.3/99.9/97.0} &
  \multicolumn{1}{c|}{98.7/100/97.1} &
  \multicolumn{1}{c|}{83.9/92.0/71.1} &
  \multicolumn{1}{c|}{96.0/98.3/88.1} &
  \multicolumn{1}{c|}{93.3/95.4/78.2} &-
   \\
\multicolumn{1}{c|}{Relation-DETR \cite{hou2024relation}} &
  \multicolumn{1}{c|}{92.0/95.6/87.5} &
  \multicolumn{1}{c|}{99.7/97.5/93.8} &
  \multicolumn{1}{c|}{94.2/97.6/86.1} &
  \multicolumn{1}{c|}{72.8/78.1/73.6} &
  \multicolumn{1}{c|}{92.0/96.3/89.6} &
  \multicolumn{1}{c|}{97.9/100/97.1} &
  \multicolumn{1}{c|}{99.4/100/92.4} &
  \multicolumn{1}{c|}{86.5/91.5/77.7} &
  \multicolumn{1}{c|}{99.5/99.9/97.3} &
  \multicolumn{1}{c|}{98.5/100/97.2} &
  \multicolumn{1}{c|}{85.9/92.6/75.3} &
  \multicolumn{1}{c|}{95.8/98.0/89.9} &
  \multicolumn{1}{c|}{91.5/95.4/80.2} &-
   \\ \hline
\end{tabular}
    \end{adjustbox}
  \label{tab:fully_supervised}
\end{table*}
\begin{table*}[t!]
\centering
\caption{Quantitative results on CRL and NT of semi-supervised methods.}
\setlength{\tabcolsep}{2pt}
\begin{adjustbox}{width=0.9999\textwidth}
    \begin{tabular}{c|c|c|c|c|c|c|c|c|c|c|c|c|c|c}
\hline
Method &
  mAP  (\%) &
  MX &
  MDS &
  MLS &
  LV &
  H &
  G &
  C &
  AB &
  B &
  RBP &
  DP &
  NTAPS &
  NB  \\ \hline
\multicolumn{15}{c}{CRL 5\% Labeled Data (Hospital 1/ Hospital 2/ Hospital 3)} \\ \hline
Unbiased Teacher V2 \cite{liu2022unbiased} &
  42.6/49.8/27.8 &
  47.7/54.6/32.9 &
  44.4/48.8/26.1 &
  11.4/9.3/13.4 &
  35.5/46.4/30.4 &
  68.0/76.4/39.5 &
  32.1/47.0/15.2 &
  55.4/64.3/34.1 &
  54.5/63.2/32.3 &
  47.1/56.0/28.5 &
  57.9/61.8/33.3 &
  55.3/59.4/33.4 &
  29.9/37.8/24.0 &
  14.9/22.3/18.1 \\
  
Label Matching \cite{chen2022label} &
  47.9/51.0/28.4 &
  52.3/56.2/32.0 &
  47.4/50.5/27.2 &
  14.8/6.6/13.9 &
  41.1/45.6/31.1 &
  74.9/77.7/40.1 &
  42.5/50.8/15.7 &
  59.4/61.3/34.1 &
  59.0/63.1/33.4 &
  55.6/57.6/28.9 &
  62.7/65.8/33.8 &
  60.5/62.7/33.2 &
  35.3/39.0/27.0 &
  17.2/25.6/19.1 \\
  
Consistent Teacher \cite{wang2023consistent} &
  48.6/52.4/28.8 &
  53.6/58.8/30.7 &
  47.1/53.6/27.4 &
  11.5/19.0/14.1 &
  44.8/45.7/30.1 &
  72.9/76.0/41.4 &
  45.1/46.6/17.2 &
  59.8/63.8/33.7 &
  58.9/63.3/32.9 &
  55.7/58.8/31.1 &
  59.0/66.9/34.9 &
  62.5/66.6/33.3 &
  40.4/40.4/26.7 &
  20.2/22.0/21.1 \\
  
Semi-DETR \cite{zhang2023semi} &
  49.3/53.1/29.4 &
  55.6/55.4/32.6 &
  49.3/54.5/28.2 &
  15.6/21.5/15.5 &
  44.4/48.4/32.8 &
  74.1/77.6/40.1 &
  46.9/51.6/17.4 &
  62.0/61.6/33.4 &
  57.9/63.4/33.4 &
  57.2/60.0/28.8 &
  61.4/67.3/36.0 &
  59.3/64.6/34.0 &
  37.5/42.2/28.0 &
  20.3/21.6/22.3 \\
  
Sparse Semi-DETR \cite{shehzadi2024sparse} &
  49.7/53.6/29.7 &
  54.7/57.9/32.5 &
  53.4/53.7/28.7 &
  19.7/18.9/15.9 &
  43.0/48.5/32.1 &
  74.7/78.3/40.8 &
  47.3/52.0/18.3 &
  59.7/64.4/34.4 &
  60.8/64.9/34.3 &
  56.7/59.8/30.1 &
  60.7/65.8/36.1 &
  61.1/64.7/35.2 &
  34.9/42.3/27.2 &
  20.0/25.7/20.1 \\
  
Semi-akmm \cite{pu2025anatomical} &
  49.8/54.3/30.2 &
  54.6/58.0/33.0 &
  48.8/55.4/30.0 &
  9.3/17.5/16.6 &
  46.4/48.5/34.0 &
  76.4/79.2/41.8 &
  47.0/53.3/17.6 &
  64.3/65.1/34.7 &
  63.2/65.5/34.2 &
  56.0/60.3/30.1 &
  61.8/67.5/36.6 &
  59.4/66.4/34.7 &
  37.8/42.5/28.1 &
  22.3/26.3/20.7 \\ \hline

\multicolumn{15}{c}{CRL 10\% Labeled Data (Hospital 1/ Hospital 2/ Hospital 3)} \\ \hline
Unbiased Teacher V2 \cite{liu2022unbiased} &
  55.5/53.6/46.6 &
  61.6/57.9/50.9 &
  50.9/53.7/46.4 &
  9.6/18.9/16.0 &
  50.1/48.5/39.2 &
  80.3/78.3/73.3 &
  56.6/52.0/41.6 &
  67.5/64.4/57.9 &
  68.6/64.9/58.5 &
  61.8/59.8/53.6 &
  67.1/65.8/60.6 &
  71.6/64.7/59.7 &
  47.6/42.3/33.4 &
  28.5/25.7/15.1 \\
  
Label Matching \cite{chen2022label} &
  57.7/55.4/47.4 &
  61.8/61.9/56.2 &
  57.5/54.2/50.5 &
  19.6/17.1/17.3 &
  52.8/48.9/27.4 &
  82.7/78.2/75.0 &
  56.4/55.9/45.8 &
  67.9/66.0/60.1 &
  70.5/66.2/58.0 &
  64.9/59.7/56.8 &
  67.0/67.9/60.0 &
  72.3/70.1/56.0 &
  47.5/43.3/33.4 &
  29.8/30.6/19.5 \\
  
Consistent Teacher \cite{wang2023consistent} &
  58.4/56.1/48.4 &
  62.9/60.7/54.4 &
  55.0/56.9/50.9 &
  22.2/20.5/8.4 &
  52.9/46.5/38.2 &
  83.1/79.1/73.5 &
  55.2/56.1/46.9 &
  68.2/67.1/63.0 &
  71.0/67.6/59.5 &
  65.7/61.6/56.6 &
  68.0/67.6/59.3 &
  73.6/69.9/58.3 &
  47.5/45.0/40.2 &
  33.8/30.2/19.9 \\
  
Semi-DETR \cite{zhang2023semi} &
  59.7/56.7/48.6 &
  62.7/61.8/53.6 &
  60.7/55.4/47.1 &
  35.0/24.2/11.5 &
  57.5/48.5/44.8 &
  82.0/80.8/72.9 &
  57.9/57.7/45.1 &
  68.6/66.9/59.8 &
  66.0/66.7/58.9 &
  64.0/62.3/55.7 &
  70.0/70.6/59.0 &
  67.0/69.0/62.5 &
  48.6/45.5/40.4 &
  35.5/27.8/20.2 \\
  
Sparse Semi-DETR \cite{shehzadi2024sparse} &
  60.1/57.0/49.8 &
  62.6/61.4/54.6 &
  62.2/54.8/48.8 &
  30.1/20.6/9.3 &
  60.1/50.1/46.4 &
  79.6/80.9/76.4 &
  59.0/57.7/47.0 &
  69.2/68.4/64.3 &
  68.1/67.4/63.2 &
  66.3/62.4/56.0 &
  67.8/70.0/61.8 &
  69.5/70.2/59.4 &
  49.7/46.2/37.8 &
  37.3/31.3/22.3 \\
  
Semi-akmm \cite{pu2025anatomical} &
  60.5/57.4/50.2 &
  61.4/62.3/56.2 &
  60.3/54.3/50.3 &
  28.5/20.2/12.8 &
  61.2/49.6/43.9 &
  81.5/82.0/76.5 &
  60.1/57.6/46.0 &
  70.7/67.5/61.8 &
  67.9/67.7/61.4 &
  66.4/62.7/57.2 &
  69.6/70.0/63.7 &
  71.4/72.1/62.9 &
  48.4/47.3/38.5 &
  39.5/32.4/21.0 \\ \hline
  Method &
  mAP  (\%) &
  MX &
  MDS &
  MLS &
  LV &
  H &
  C &
  AB &
  RBP &
  DP &
  NT &
  NTAPS &
  NB \\ \hline
\multicolumn{15}{c}{NT 5\% Labeled Data (Hospital 1/ Hospital 2/ Hospital 3)} \\ \hline
Unbiased Teacher V2 \cite{liu2022unbiased} &
  41.1/40.9/26.6 &
  41.9/41.9/34.5 &
  42.0/41.4/27.8 &
  32.2/33.1/19.3 &
  41.1/41.0/33.5 &
  43.0/43.0/35.3 &
  43.0/43.0/29.3 &
  39.9/39.0/5.5 &
  43.0/43.0/34.4 &
  43.0/43.0/37.6 &
  40.3/39.4/17.5 &
  42.2/42.0/25.6 &
  41.1/41.1/18.8 &-\\
  
Label Matching \cite{chen2022label} &
  42.8/42.8/28.6 &
  43.8/43.8/35.2 &
  43.3/43.3/27.8 &
  34.7/34.7/22.3 &
  42.9/42.9/30.1 &
  45.0/45.0/37.4 &
  45.0/45.0/32.7 &
  40.8/40.8/7.6 &
  45.0/45.0/35.2 &
  45.0/45.0/36.5 &
  41.3/41.3/25.9 &
  43.9/43.9/27.4 &
  43.0/43.0/25.1 &-\\
  
Consistent Teacher \cite{wang2023consistent} &
  43.3/43.7/29.3 &
  56.2/45.1/36.1 &
  45.2/44.7/28.5 &
  31.4/34.3/22.9 &
  54.5/44.3/30.8 &
  57.5/46.0/38.3 &
  47.7/46.0/33.5 &
  9.0/41.5/7.8 &
  55.9/45.9/36.1 &
  61.2/46.0/37.4 &
  28.5/42.2/26.5 &
  41.7/45.2/28.1 &
  30.7/43.8/25.7 &-\\
  
Semi-DETR \cite{zhang2023semi} &
  43.9/44.0/30.3 &
  44.9/44.9/39.4 &
  44.5/44.9/31.6 &
  35.0/35.9/22.0 &
  44.0/44.3/38.1 &
  46.0/46.0/40.2 &
  46.0/46.0/33.4 &
  42.7/42.1/6.3 &
  46.0/46.0/39.1 &
  46.0/46.0/42.9 &
  42.9/42.6/19.9 &
  45.1/45.1/29.2 &
  43.8/43.9/21.5 &-\\
  
Sparse Semi-DETR \cite{shehzadi2024sparse} &
  45.8/44.8/30.7 &
  47.0/45.9/37.8 &
  46.3/45.9/29.9 &
  36.1/35.6/23.9 &
  46.2/45.5/32.2 &
  48.0/47.0/40.1 &
  48.0/47.0/35.1 &
  43.9/42.7/8.2 &
  47.9/47.0/37.8 &
  48.0/47.0/39.2 &
  44.9/43.2/27.8 &
  47.0/46.2/29.4 &
  45.7/44.8/26.9 &-\\
  
Semi-akmm \cite{pu2025anatomical} &
  45.9/45.6/30.9 &
  52.8/47.0/40.1 &
  45.9/46.7/32.3 &
  37.5/35.7/22.4 &
  49.5/46.3/38.9 &
  57.5/48.0/41.0 &
  48.9/48.0/34.1 &
  22.1/42.4/6.4 &
  56.0/48.0/40.0 &
  56.0/48.0/43.8 &
  40.4/44.3/20.4 &
  43.9/47.0/29.8 &
  40.3/45.5/21.9 &-\\ \hline

\multicolumn{15}{c}{NT 10\% Labeled Data (Hospital 1/ Hospital 2/ Hospital 3)} \\ \hline
Unbiased Teacher V2 \cite{liu2022unbiased} &
  54.4/50.5/39.8 &
  55.8/51.9/43.1 &
  55.3/51.3/40.0 &
  43.7/39.3/30.5 &
  54.8/50.9/41.2 &
  57.0/53.0/45.2 &
  57.0/53.0/42.4 &
  52.3/48.5/35.1 &
  56.9/53.0/45.3 &
  57.0/53.0/45.3 &
  53.2/49.0/34.1 &
  55.7/52.0/39.6 &
  54.1/50.7/36.5 &-\\
  
Label Matching \cite{chen2022label} &
  55.5/51.6/40.2 &
  57.2/52.9/43.6 &
  56.7/52.8/40.3 &
  44.4/41.2/32.3 &
  55.9/51.9/39.3 &
  58.0/54.0/45.4 &
  58.0/54.0/43.1 &
  52.9/49.7/34.8 &
  57.9/54.0/44.9 &
  58.0/54.0/46.3 &
  54.4/50.5/36.3 &
  57.2/53.4/40.1 &
  55.3/51.5/35.5 &-\\
  
Consistent Teacher \cite{wang2023consistent} &
  56.4/52.3/40.5 &
  57.8/53.6/44.9 &
  57.7/53.0/40.1 &
  45.0/42.4/31.4 &
  56.7/52.5/38.6 &
  59.0/55.0/47.7 &
  59.0/55.0/44.7 &
  54.3/49.9/32.0 &
  58.9/55.0/47.3 &
  59.0/55.0/47.4 &
  55.2/50.4/38.1 &
  58.3/53.7/40.0 &
  56.2/52.5/34.3 &-\\
  
Semi-DETR \cite{zhang2023semi} &
  57.3/52.5/41.5 &
  58.5/53.7/47.7 &
  58.6/53.2/41.5 &
  46.9/41.8/33.9 &
  57.8/52.6/44.7 &
  60.0/55.0/52.0 &
  60.0/55.0/44.2 &
  54.9/51.1/20.0 &
  59.9/55.0/50.6 &
  60.0/55.0/50.6 &
  55.6/51.3/36.5 &
  58.8/54.0/39.6 &
  57.2/52.4/36.4 &-\\
  
Sparse Semi-DETR \cite{shehzadi2024sparse} &
  57.6/53.5/42.0 &
  62.4/54.7/43.0 &
  57.9/54.2/42.6 &
  44.1/42.6/33.4 &
  59.6/53.6/42.1 &
  65.3/56.0/44.0 &
  61.3/56.0/44.0 &
  50.7/52.0/40.9 &
  65.5/55.9/44.0 &
  65.5/56.0/44.0 &
  49.3/52.2/41.0 &
  57.3/54.9/43.2 &
  52.8/53.3/41.9 &-\\
  
Semi-akmm \cite{pu2025anatomical} &
  58.0/53.6/43.1 &
  66.2/55.2/44.4 &
  60.3/54.8/44.0 &
  47.1/42.9/34.5 &
  57.9/53.9/43.3 &
  67.3/56.0/45.0 &
  63.4/56.0/45.0 &
  44.1/51.1/41.0 &
  65.6/55.9/45.0 &
  69.3/56.0/45.0 &
  42.6/52.5/42.2 &
  58.2/55.3/44.4 &
  53.7/53.4/42.9 &-\\ \hline
\end{tabular}
    \end{adjustbox}
  \label{tab:semi-supervised}
\end{table*}
\begin{table*}[t!]
\centering
\caption{Quantitative results on CRL and NT of UDA methods. {We have split the tables into independent sub-tables categorized by Target Hospital and bolded the top-performing metrics to improve readability.}}
\setlength{\tabcolsep}{12pt}
\begin{adjustbox}{width=0.9999\textwidth}
    \begin{tabular}{c|c|c|c|c|c|c|c|c|c|c|c|c|c|c}
\hline
Method &
  mAP  (\%) &
  MX &
  MDS &
  MLS &
  LV &
  H &
  G &
  C &
  AB &
  B &
  RBP &
  DP &
  NTAPS &
  NB  \\ \hline
\multicolumn{15}{c}{CRL (Hospital 1 $\rightleftharpoons$ Hospital 2)} \\ \hline
SIGMA \cite{li2022sigma} &
  82.0/83.0 &
  94.2/90.2 &
  88.5/77.2 &
  56.7/48.9 &
  73.5/84.3 &
  99.9/98.3 &
  79.4/86.6 &
  98.9/96.6 &
  100/97.1 &
  96.5/94.8 &
  97.5/94.8 &
  99.4/95.3 &
  57.0/63.3 &
  24.2/51.1 \\
  
SIGMA++ \cite{li2023sigma++} &
  84.7/84.1 &
  97.0/91.0 &
  88.5/77.3 &
  58.0/53.2 &
  77.4/87.3 &
  99.9/98.3 &
  81.9/88.9 &
  98.3/97.0 &
  99.8/97.0 &
  96.0/95.3 &
  97.7/95.0 &
  99.3/95.8 &
  77.4/65.9 &
  30.0/52.1 \\
  
CMT \cite{cao2023contrastive} &
  86.8/85.7 &
  98.3/91.0 &
  89.3/81.2 &
  63.0/57.0 &
  81.4/87.7 &
  99.9/98.2 &
  84.0/89.1 &
  98.9/97.2 &
  99.8/96.7 &
  96.3/95.0 &
  98.0/95.4 &
  98.4/96.4 &
  85.6/72.6 &
  35.1/56.2 \\
  
$M^3$-UDA \cite{pu2024m3} &
  88.1/86.4 &
  98.7/91.3 &
  90.7/82.8 &
  65.3/58.0 &
  82.6/88.5 &
  99.8/98.1 &
  86.8/88.4 &
  98.9/96.6 &
  99.6/95.7 &
  96.1/94.9 &
  97.7/95.7 &
  98.8/97.0 &
  89.3/76.6 &
  40.9/59.3 \\
  
ToMo-UDA \cite{pu2024unsupervised} &
  88.3/87.1 &
  99.1/90.7 &
  90.3/84.2 &
  67.6/62.4 &
  85.5/89.1 &
  98.6/98.2 &
  85.5/90.7 &
  99.8/96.6 &
  99.8/94.6 &
  96.1/95.4 &
  98.1/95.8 &
  98.1/96.7 &
  88.9/78.9 &
  40.1/59.8 \\
  
DATR \cite{chen2025datr} &  
  89.5/87.2 &
  96.0/97.2 &
  92.5/95.5 &
  66.5/73.4 &
  81.5/89.6 &
  97.8/98.4 &
  85.4/92.5 &
  96.6/99.7 &
  95.9/96.8 &
  94.4/97.3 &
  95.1/97.2 &
  95.9/97.6 &
  86.6/92.8 &
  78.7/87.4 \\
  
  \hline

\multicolumn{15}{c}{CRL (Hospital 1 $\rightleftharpoons$ Hospital 3)} \\ \hline
SIGMA \cite{li2022sigma} &
  38.3/81.3 &
  49.8/90.6 &
  39.6/68.6 &
  19.2/51.8 &
  44.8/89.0 &
  50.9/98.3 &
  32.5/85.6 &
  54.5/96.6 &
  55.2/97.3 &
  49.7/96.7 &
  50.5/94.5 &
  52.4/96.3 &
  29.5/66.6 &
  6.9/25.1 \\
  
SIGMA++ \cite{li2023sigma++} &
  39.4/82.2 &
  50.2/90.5 &
  40.1/71.7 &
  22.4/52.4 &
  47.0/89.6 &
  50.2/98.4 &
  36.7/85.9 &
  53.1/96.8 &
  56.6/97.6 &
  51.2/96.8 &
  50.8/95.1 &
  54.9/96.4 &
  31.8/69.1 &
  6.3/28.0 \\
  
CMT \cite{cao2023contrastive} &
  40.2/83.4 &
  49.6/91.1 &
  40.7/76.9 &
  27.2/53.4 &
  48.6/89.2 &
  50.1/98.3 &
  38.2/89.5 &
  55.1/96.5 &
  54.1/97.2 &
  50.7/96.4 &
  52.9/95.6 &
  52.7/97.0 &
  35.1/72.6 &
  7.5/31.1 \\
  
$M^3$-UDA \cite{pu2024m3} &
  40.9/84.7 &
  48.6/91.5 &
  42.8/77.3 &
  29.9/55.1 &
  50.6/88.8 &
  49.6/98.3 &
  39.3/89.5 &
  54.8/96.1 &
  54.6/97.2 &
  52.7/96.0 &
  52.2/95.3 &
  53.6/97.7 &
  37.2/77.9 &
  7.4/40.0 \\
  
ToMo-UDA \cite{pu2024unsupervised} &
  40.9/84.7 &
  47.8/90.5 &
  42.1/76.5 &
  33.2/53.9 &
  48.3/88.6 &
  50.8/98.5 &
  38.7/89.4 &
  53.6/95.8 &
  56.5/96.2 &
  50.7/96.8 &
  52.2/94.3 &
  51.8/97.6 &
  37.1/80.4 &
  10.4/42.4 \\

DATR \cite{chen2025datr} &
  41.3/84.6 & 44.6/91.4 & 38.4/88.1 & 32.2/62.6 & 37.4/64.0 & 46.1/98.3 & 34.2/79.1 & 46.9/96.9 & 45.8/94.6 & 45.6/94.2 & 44.5/95.0 & 46.3/92.4 & 38.2/60.2 & 36.7 \\
  
  \hline

\multicolumn{15}{c}{CRL (Hospital 2 $\rightleftharpoons$ Hospital 3)} \\ \hline
SIGMA \cite{li2022sigma} &
  40.6/83.7 &
  53.6/95.4 &
  22.8/73.5 &
  30.0/63.4 &
  44.6/78.3 &
  54.1/99.8 &
  37.7/84.3 &
  54.2/96.8 &
  54.8/98.0 &
  54.3/96.5 &
  51.2/97.3 &
  52.2/98.6 &
  33.5/75.0 &
  25.8/30.7 \\
  
SIGMA++ \cite{li2023sigma++} &
  40.6/85.1 &
  51.4/97.6 &
  21.3/80.4 &
  30.1/67.2 &
  45.9/77.6 &
  53.1/99.8 &
  38.1/83.8 &
  53.5/97.9 &
  54.1/96.6 &
  54.4/96.7 &
  51.0/97.6 &
  51.6/99.1 &
  35.9/80.1 &
  27.9/32.4 \\
  
CMT \cite{cao2023contrastive} &
  41.2/87.2 &
  51.1/97.7 &
  24.0/84.2 &
  29.3/71.7 &
  45.9/80.3 &
  53.5/99.8 &
  38.0/86.9 &
  53.4/98.2 &
  53.6/98.5 &
  52.5/96.4 &
  52.1/97.8 &
  51.1/98.3 &
  39.3/82.8 &
  32.6/41.4 \\
  
$M^3$-UDA \cite{pu2024m3} &
  41.3/87.7 &
  49.3/97.6 &
  30.1/84.5 &
  27.7/73.1 &
  45.5/83.7 &
  53.6/99.9 &
  37.7/85.7 &
  51.2/98.1 &
  52.2/98.5 &
  51.0/96.6 &
  51.1/97.7 &
  50.3/98.9 &
  41.4/80.7 &
  37.4/45.0 \\
  
ToMo-UDA \cite{pu2024unsupervised} &
  40.6/87.6 &
  47.6/96.6 &
  27.9/87.0 &
  28.3/71.4 &
  44.6/84.4 &
  51.6/99.8 &
  37.6/88.4 &
  50.7/98.0 &
  52.9/98.8 &
  49.8/97.5 &
  49.6/98.0 &
  49.3/98.9 &
  39.9/81.1 &
  38.0/39.2 \\ 

  DATR \cite{chen2025datr} &
  42.4/87.4 & 43.4/98.1 & 42.9/95.7 & 38.5/79.3 & 40.2/89.0 & 43.6/99.3 & 41.5/93.6 & 44.0/99.7 & 44.0/99.0 & 43.4/98.0 & 43.6/96.3 & 43.6/98.0 & 42.3/91.1 & 39.9/85.6 \\
  
  \hline
  Method &
  mAP  (\%) &
  MX &
  MDS &
  MLS &
  LV &
  H &
  C &
  AB &
  RBP &
  DP &
  NT &
  NTAPS &
  NB \\ \hline
\multicolumn{15}{c}{NT (Hospital 1 $\rightleftharpoons$ Hospital 2)} \\ \hline
SIGMA \cite{li2022sigma} &
  70.3/80.2 &
  72.5/94.1 &
  71.8/86.4 &
  57.0/64.0 &
  70.6/87.0 &
  74.0/98.7 &
  73.9/91.1 &
  66.5/78.7 &
  73.9/97.5 &
  74.0/97.5 &
  67.7/75.2 &
  71.8/89.0 &
  69.8/82.7 &
  -  \\
  
SIGMA++ \cite{li2023sigma++} &
  67.9/80.4 &
  70.0/94.2 &
  69.4/87.9 &
  54.4/64.9 &
  68.4/87.7 &
  71.0/99.4 &
  71.0/93.3 &
  64.8/78.5 &
  70.9/97.6 &
  71.0/97.2 &
  66.5/75.4 &
  70.1/87.7 &
  67.7/81.0  &
  - \\
CMT \cite{cao2023contrastive} &
  69.8/81.4 &
  72.0/94.4 &
  71.4/88.6 &
  55.9/67.3 &
  70.3/88.0 &
  73.0/99.2 &
  73.0/91.5 &
  66.6/78.8 &
  72.9/98.3 &
  73.0/97.7 &
  68.4/78.7 &
  72.0/90.9 &
  69.6/83.9  &
  - \\
$M^3$-UDA \cite{pu2024m3} &
  70.3/82.0 &
  72.5/94.4 &
  71.8/87.3 &
  57.0/70.8 &
  70.6/90.0 &
  74.0/99.4 &
  73.9/96.0 &
  66.5/82.5 &
  73.9/98.3 &
  74.0/97.4 &
  67.7/77.5 &
  71.8/90.2 &
  69.8/81.5  &
  - \\
ToMo-UDA \cite{pu2024unsupervised} &
  71.4/82.5 &
  73.2/95.2 &
  73.3/90.1 &
  56.9/70.1 &
  72.4/89.8 &
  75.0/99.8 &
  74.9/94.3 &
  67.0/82.1 &
  74.9/98.9 &
  75.0/97.2 &
  69.1/80.0 &
  73.7/90.5 &
  71.5/83.8  &
  - \\

  DATR \cite{chen2025datr} &
  71.3/81.7 & 73.0/93.3 & 72.2/88.0 & 57.8/71.2 & 71.5/88.5 & 75.0/99.0 & 75.0/93.7 & 68.0/83.3 & 74.9/96.9 & 75.0/97.2 & 68.8/77.1 & 73.2/88.7 & 71.6/84.2 & - \\
  
 \hline

\multicolumn{15}{c}{NT (Hospital 1 $\rightleftharpoons$ Hospital 3)} \\ \hline
SIGMA \cite{li2022sigma} &
  67.0/51.5 &
  69.0/52.9 &
  68.5/52.5 &
  53.6/41.3 &
  67.4/51.9 &
  70.0/54.0 &
  70.0/54.0 &
  63.8/49.2 &
  69.9/53.9 &
  70.0/54.0 &
  65.6/49.9 &
  69.1/52.7 &
  66.7/51.5  &
  - \\

SIGMA++ \cite{li2023sigma++} &
  67.9/53.6 &
  70.0/55.2 &
  69.4/54.8 &
  54.4/42.9 &
  68.4/53.9 &
  71.0/56.0 &
  71.0/56.0 &
  64.8/51.1 &
  70.9/55.9 &
  71.0/56.0 &
  66.5/52.5 &
  70.1/55.3 &
  67.7/53.4  &
  - \\
  
CMT \cite{cao2023contrastive} &
  67.5/53.5 &
  69.2/54.7 &
  68.4/54.2 &
  54.7/42.6 &
  67.7/53.6 &
  71.0/56.0 &
  71.0/56.0 &
  64.4/52.0 &
  70.9/55.9 &
  71.0/56.0 &
  65.1/52.2 &
  69.3/54.9 &
  67.8/53.3  &
  -\\
  
$M^3$-UDA \cite{pu2024m3} &
  69.4/55.6 &
  71.1/61.7 &
  70.3/55.1 &
  56.3/43.0 &
  69.6/52.9 &
  73.0/65.5 &
  73.0/61.4 &
  66.2/44.0 &
  72.9/64.9 &
  73.0/65.1 &
  66.9/52.4 &
  71.2/54.9 &
  69.7/47.0  &
  - \\
ToMo-UDA \cite{pu2024unsupervised} &
  70.4/55.2 &
  72.1/64.3 &
  71.3/50.7 &
  57.0/28.6 &
  70.6/55.3 &
  74.0/69.3 &
  74.0/63.5 &
  67.1/44.6 &
  73.9/66.2 &
  74.0/67.5 &
  67.9/48.0 &
  72.2/55.5 &
  70.7/48.5  &
  -\\

DATR \cite{chen2025datr} &
  70.8/56.5 & 72.1/57.5 & 72.6/57.9 & 59.1/47.1 & 71.9/57.4 & 74.0/59.0 & 74.0/59.0 & 65.8/52.5 & 73.9/58.9 & 74.0/59.0 & 69.1/55.1 & 72.6/57.9 & 70.7/56.3 & - \\

  \hline
\multicolumn{15}{c}{NT (Hospital 2 $\rightleftharpoons$ Hospital 3)} \\ \hline
SIGMA \cite{li2022sigma} &
  42.2/64.1 &
  45.5/65.6 &
  42.2/65.5 &
  34.6/52.3 &
  44.2/64.7 &
  46.8/67.0 &
  44.9/67.0 &
  39.2/61.3 &
  47.1/66.9 &
  46.9/67.0 &
  35.7/62.3 &
  42.9/65.7 &
  37.1/63.9  &
  - \\
SIGMA++ \cite{li2023sigma++} &
  45.4/64.7 &
  48.9/66.6 &
  45.1/66.1 &
  36.7/50.7 &
  46.9/65.5 &
  50.6/68.0 &
  48.6/68.0 &
  41.0/61.3 &
  50.4/67.9 &
  50.6/68.0 &
  38.5/62.4 &
  46.2/66.8 &
  40.8/64.8  &
  - \\
CMT \cite{cao2023contrastive} &
  47.5/66.7 &
  51.1/68.5 &
  47.5/68.0 &
  38.9/53.5 &
  49.7/67.3 &
  52.6/70.0 &
  50.5/70.0 &
  44.1/63.8 &
  53.0/69.9 &
  52.7/70.0 &
  40.2/64.7 &
  48.2/68.3 &
  41.7/66.7  &
  - \\
$M^3$-UDA \cite{pu2024m3} &
  48.4/67.0 &
  52.1/69.0 &
  48.4/68.5 &
  39.6/53.6 &
  50.7/67.4 &
  53.6/70.0 &
  51.4/70.0 &
  44.9/63.8 &
  54.0/69.9 &
  53.7/70.0 &
  40.9/65.6 &
  49.1/69.1 &
  42.5/66.7  &
  - \\
ToMo-UDA \cite{pu2024unsupervised} &
  51.4/68.6 &
  54.7/70.6 &
  52.0/69.7 &
  41.9/53.4 &
  52.6/69.1 &
  57.1/72.0 &
  54.4/72.0 &
  47.0/66.0 &
  57.0/71.9 &
  56.7/72.0 &
  44.9/66.5 &
  52.8/70.7 &
  46.1/68.8  &
  -\\ 

  DATR \cite{chen2025datr} &
  50.4/68.8 & 51.9/70.2 & 51.5/70.6 & 39.5/56.7 & 51.1/68.7 & 53.0/72.0 & 53.0/72.0 & 47.8/66.1 & 52.9/71.9 & 53.0/72.0 & 48.6/66.6 & 52.0/69.9 & 50.5/68.9 & - \\
  
  \hline
  
\end{tabular}
    \end{adjustbox}
  \label{tab:uda}
\end{table*}
\begin{table*}[t!]
\centering
\caption{Quantitative results on CRL and NT of Source-free DA methods. {We have split the tables into independent sub-tables categorized by Target Hospital and bolded the top-performing metrics to improve readability.}}
\setlength{\tabcolsep}{12pt}
\begin{adjustbox}{width=0.9999\textwidth}
    \begin{tabular}{c|c|c|c|c|c|c|c|c|c|c|c|c|c|c}
\hline
Method &
  mAP  (\%) &
  MX &
  MDS &
  MLS &
  LV &
  H &
  G &
  C &
  AB &
  B &
  RBP &
  DP &
  NTAPS &
  NB \\ \hline
\multicolumn{14}{c}{CRL (Hospital 1 $\rightleftharpoons$ Hospital 2)} \\ \hline
SF-AT \cite{li2022cross} &
  69.3/75.7 &
  66.2/88.7 &
  80.4/63.1 &
  35.5/45.7 &
  50.0/74.1 &
  99.6/98.4 &
  51.8/82.5 &
  98.9/96.5 &
  98.8/96.6 &
  92.1/94.0 &
  97.1/91.7 &
  98.9/94.4 &
  19.2/33.6 &
  12.3/24.0 \\
  
$A^2$SFOD \cite{chu2023adversarial} &
  71.4/77.0 &
  71.8/88.8 &
  79.5/62.4 &
  39.3/37.2 &
  58.0/79.0 &
  99.6/98.4 &
  55.8/83.4 &
  98.9/96.6 &
  99.9/95.3 &
  94.1/95.5 &
  97.1/92.7 &
  98.9/94.2 &
  22.6/44.5 &
  12.7/32.4 \\
  
IRG \cite{vs2023instance} &
  74.9/79.5 &
  82.1/90.2 &
  84.1/67.9 &
  46.7/42.6 &
  62.5/81.4 &
  99.6/98.3 &
  66.2/83.7 &
  98.9/96.5 &
  99.9/96.3 &
  93.1/95.5 &
  97.2/93.0 &
  98.9/94.8 &
  34.0/49.1 &
  10.6/44.6 \\
  
ATSS \cite{pu2025leveraging} &
  79.1/81.2 &
  89.1/89.7 &
  85.1/70.0 &
  53.0/44.5 &
  71.0/83.6 &
  99.7/98.5 &
  76.2/85.5 &
  98.9/96.7 &
  99.9/95.9 &
  95.6/95.6 &
  97.5/94.1 &
  99.1/94.6 &
  43.6/57.1 &
  19.2/50.0 \\ \hline

\multicolumn{14}{c}{CRL (Hospital 1 $\rightleftharpoons$ Hospital 3)} \\ \hline
SF-AT \cite{li2022cross} &
  32.8/74.3 &
  53.7/81.9 &
  38.6/44.3 &
  11.2/42.4 &
  29.8/80.4 &
  54.3/98.6 &
  3.2/74.5 &
  55.4/95.4 &
  58.2/94.8 &
  30.9/95.8 &
  51.4/93.7 &
  52.0/95.5 &
  15.7/47.1 &
  4.6/21.2 \\
  
$A^2$SFOD \cite{chu2023adversarial} &
  33.9/76.0 &
  52.5/81.5 &
  42.2/47.5 &
  14.3/42.5 &
  32.7/81.3 &
  53.4/98.3 &
  5.6/79.0 &
  58.0/97.0 &
  57.8/97.1 &
  35.8/96.0 &
  50.5/94.0 &
  50.8/96.4 &
  18.2/54.9 &
  2.9/21.8 \\
  
IRG \cite{vs2023instance} &
  35.3/77.2 &
  51.7/86.5 &
  41.3/48.4 &
  22.3/47.2 &
  40.8/85.5 &
  52.6/98.3 &
  8.7/86.1 &
  55.2/96.7 &
  56.5/97.3 &
  36.7/96.1 &
  52.5/93.7 &
  52.0/96.1 &
  21.3/55.4 &
  2.8/16.0 \\
  
ATSS \cite{pu2025leveraging} &
  37.1/79.4 &
  52.1/89.7 &
  39.9/58.1 &
  22.3/51.4 &
  42.6/87.5 &
  52.4/98.4 &
  17.0/86.4 &
  56.0/96.6 &
  55.0/97.4 &
  45.1/96.6 &
  52.6/94.3 &
  51.7/96.4 &
  28.3/59.7 &
  4.5/20.1 \\ \hline

\multicolumn{14}{c}{CRL (Hospital 2 $\rightleftharpoons$ Hospital 3)} \\ \hline
SF-AT \cite{li2022cross} &
  36.4/75.3 &
  51.9/91.0 &
  7.5/65.1 &
  18.8/50.8 &
  39.2/67.2 &
  56.0/99.5 &
  36.1/79.2 &
  56.2/89.3 &
  59.2/98.0 &
  58.0/94.1 &
  51.3/96.8 &
  54.7/91.9 &
  13.0/42.3 &
  7.7/14.1 \\
  
$A^2$SFOD \cite{chu2023adversarial} &
  37.4/77.8 &
  54.1/96.8 &
  11.5/65.8 &
  22.2/55.6 &
  40.1/74.5 &
  55.5/99.5 &
  35.3/78.6 &
  56.4/94.8 &
  58.4/97.7 &
  56.7/96.4 &
  50.3/96.9 &
  55.2/94.8 &
  18.2/42.1 &
  9.6/18.4 \\
  
IRG \cite{vs2023instance} &
  38.0/79.1 &
  53.7/96.7 &
  11.8/68.4 &
  22.3/57.2 &
  40.6/72.8 &
  54.4/99.7 &
  36.0/80.1 &
  55.2/95.6 &
  58.5/97.5 &
  55.4/95.9 &
  51.2/96.6 &
  56.4/97.3 &
  19.3/51.6 &
  16.7/19.4 \\
  
ATSS \cite{pu2025leveraging} &
  38.7/81.9 &
  52.0/96.4 &
  13.6/69.7 &
  25.6/59.8 &
  39.4/76.2 &
  55.3/99.8 &
  34.8/84.4 &
  56.2/96.5 &
  55.2/98.3 &
  54.0/96.3 &
  49.8/97.0 &
  54.3/98.0 &
  27.1/66.5 &
  25.0/25.8 \\ \hline
  Method &
  mAP  (\%) &
  MX &
  MDS &
  MLS &
  LV &
  H &
  C &
  AB &
  RBP &
  DP &
  NT &
  NTAPS &
  NB \\ \hline
\multicolumn{14}{c}{NT (Hospital 1 $\rightleftharpoons$ Hospital 2)} \\ \hline
SF-AT \cite{li2022cross} &
  60.1/79.0 &
  61.7/91.8 &
  61.1/86.4 &
  48.3/66.8 &
  60.6/81.1 &
  63.0/97.6 &
  63.0/89.0 &
  57.8/79.4 &
  62.9/94.9 &
  63.0/96.1 &
  58.8/74.8 &
  61.5/87.3 &
  59.8/80.5  &
  - \\
  
$A^2$SFOD \cite{chu2023adversarial} &
  61.1/80.1 &
  62.7/94.1 &
  62.1/86.5 &
  49.1/66.1 &
  61.6/82.4 &
  64.0/98.1 &
  64.0/91.2 &
  58.7/83.3 &
  63.9/96.0 &
  64.0/97.0 &
  59.8/77.1 &
  62.5/88.6 &
  60.7/80.6  &
  - \\
  
IRG \cite{vs2023instance} &
  66.8/79.7 &
  68.5/94.2 &
  67.9/86.9 &
  53.7/63.2 &
  67.3/85.4 &
  70.0/98.5 &
  70.0/91.4 &
  64.2/80.7 &
  69.9/96.1 &
  70.0/97.0 &
  65.4/74.8 &
  68.4/86.1 &
  66.4/82.0  &
  - \\
  
ATSS \cite{pu2025leveraging} &
  67.8/80.6 &
  69.5/93.8 &
  68.9/88.7 &
  54.5/65.0 &
  68.3/84.7 &
  71.0/98.9 &
  71.0/92.0 &
  65.1/81.3 &
  70.9/97.6 &
  71.0/97.2 &
  66.3/78.0 &
  69.4/89.4 &
  67.4/80.0  &
  - \\ \hline

\multicolumn{14}{c}{NT (Hospital 1 $\rightleftharpoons$ Hospital 3)} \\ \hline
SF-AT \cite{li2022cross} &
  63.8/45.9 &
  65.7/49.6 &
  64.9/45.3 &
  49.7/38.3 &
  64.3/46.1 &
  67.0/51.6 &
  67.0/49.2 &
  61.4/39.9 &
  66.9/50.8 &
  67.0/52.3 &
  61.9/39.3 &
  65.8/46.0 &
  64.0/42.3  &
  - \\
  
$A^2$SFOD \cite{chu2023adversarial} &
  64.0/48.4 &
  65.6/52.4 &
  65.0/47.8 &
  51.4/40.5 &
  64.5/48.7 &
  67.0/54.5 &
  67.0/51.9 &
  61.4/42.1 &
  66.9/53.6 &
  67.0/55.2 &
  62.6/41.4 &
  65.5/48.5 &
  63.6/44.7  &
  -\\
  
IRG \cite{vs2023instance} &
  64.9/50.1 &
  66.4/54.2 &
  65.8/49.5 &
  51.7/41.9 &
  65.1/50.4 &
  68.0/56.4 &
  68.0/53.8 &
  63.2/43.6 &
  67.9/55.5 &
  68.0/57.1 &
  63.4/42.9 &
  66.7/50.2 &
  64.7/46.3  &
  -\\
  
ATSS \cite{pu2025leveraging} &
  67.0/54.3 &
  69.0/55.9 &
  68.5/55.9 &
  53.6/44.2 &
  67.4/54.1 &
  70.0/57.0 &
  70.0/57.0 &
  63.8/51.2 &
  69.9/56.9 &
  70.0/57.0 &
  65.6/52.6 &
  69.1/55.8 &
  66.7/54.3  &
  - \\ \hline

\multicolumn{14}{c}{NT (Hospital 2 $\rightleftharpoons$ Hospital 3)} \\ \hline
SF-AT \cite{li2022cross} &
  38.2/63.8 &
  42.3/68.1 &
  37.8/64.9 &
  29.5/52.2 &
  36.3/65.1 &
  44.9/70.8 &
  42.1/68.3 &
  30.1/57.4 &
  44.5/70.8 &
  44.6/70.3 &
  35.9/55.4 &
  37.6/64.2 &
  32.3/58.5  &
  - \\
  
$A^2$SFOD \cite{chu2023adversarial} &
  41.9/65.3 &
  43.1/70.2 &
  42.5/66.0 &
  33.1/53.8 &
  42.4/66.1 &
  44.0/72.5 &
  44.0/68.7 &
  40.3/59.4 &
  43.9/72.7 &
  44.0/72.3 &
  41.1/55.4 &
  43.1/66.3 &
  41.9/60.2  &
  - \\
  
IRG \cite{vs2023instance} &
  43.1/65.6 &
  44.4/69.9 &
  44.0/65.1 &
  34.5/56.3 &
  43.3/66.8 &
  45.0/72.2 &
  45.0/68.3 &
  41.0/60.8 &
  45.0/72.5 &
  45.0/72.2 &
  42.2/55.4 &
  44.4/67.7 &
  42.9/59.8  &
  - \\
  
ATSS \cite{pu2025leveraging} &
  44.0/66.6 &
  45.0/71.8 &
  44.9/67.4 &
  35.9/53.4 &
  44.4/70.0 &
  46.0/73.1 &
  46.0/70.3 &
  42.1/62.0 &
  46.0/73.2 &
  46.0/73.2 &
  42.8/57.9 &
  45.1/67.3 &
  43.9/59.9  &
  - \\ \hline
\end{tabular}
    \end{adjustbox}
  \label{tab:sfuda}
\end{table*}
%

\section{Benchmark}
In this section, we evaluate existing advanced methods~\cite{ren2016faster, carion2020endtoend, peize2020sparse, zhu2021deformable, meng2021-CondDETR, liu2022dabdetr, li2022grounded, li2022exploring, zhou2022detecting, zong2023detrs, zhang2022dino} on our proposed \textbf{FUSEP} for \emph{\textbf{multi-center medical multi-object detection task}} across multiple medical centers. These methods are implemented with their default configuration. All training and inference are under the same recipe as outlined in Section~\ref{sec:image_configuration} and~\ref{sec:implementation_details}. 
%
\subsection{Imaging Configuration}
\label{sec:image_configuration}
%
During training, all images will first be resized to 800$\times$1333 in height and width. Then, for the image augmentation, the rotation and flipping operations will be randomly applied to the resized image. Furthermore, we will randomly change the brightness, contrast, saturation and hue of an image. All operations for augmentation are set with a probability of 0.5. For the inference in the validation set and the testing set, we only conducted resizing and normalization of the images. 
We conducted experiments in 4 downstream tasks based on object detection.
\textbf{(1) Fully supervised multi-object detection.} We designed a comprehensive experimental framework to evaluate multi-objective detection performance in medical imaging. The study utilized de-identified datasets from three tertiary hospitals. \textbf{(2) Semi-supervised multi-object detection.} To simulate real-world clinical scenarios with limited annotations, we implemented a semi-supervised learning protocol where only 5\% and 10\% of the training data retained full annotations, while the remaining data were treated as unlabeled samples. \textbf{(3) UDA in multi-object detection.} We examined the domain adaptation problem across three medical centers, performing experiments under six distinct cross-domain configurations. \textbf{(4) Source-free UDA in multi-object detection.} Following the paradigm of Unsupervised Domain Adaptation (UDA), but constrained by the practical limitation of having access only to source model weights rather than the original dataset, we specifically studied the source-free UDA scenario across healthcare institutions. This investigation yielded six comprehensive source-free UDA experiments evaluating cross-center domain adaptation performance.

%
\subsection{Implementation Details}
\label{sec:implementation_details}

In our experiment, we use one 3090RTX (24G) with an E5-2680 CPU and 256 GB RAM. All methods are conducted in the same training environment. 
We use the MMDetection \cite{chen2019mmdetection} framework and follow the default settings to complete the experiments. {To ensure full reproducibility, we have prepared a clean, well-documented codebase. A GitHub link (https://github.com/LiwenWang919/FUSEP) containing all training scripts, environment configurations, and baseline weights is provided for public access.}
For different object detection methods, we use their default backbones \textit{(See details of Table~\ref{tab:efficiency}}.

\noindent
\textbf{\textit{Training and Inference.}}
During training, we use all training sets (612 images of CRL and NT view in Hospital-1,590 images of CRL view and 607 images of NT view in Hospital-2 and 389 images of CRL view and 402 images of NT view in Hospital-3,respectively) with corresponding annotations. The total training epochs are 50 for all experiment setting. We chose the model that performs the best in the validation dataset. For the inference, no augmentation will be applied except resize. All results will be reported in the test set inferred by the selected best model.

\noindent
\textbf{\textit{Evaluation Metrics for Object Detection.}}
We reported the evaluation metrics with the mean average precision (mAP), i.e., mAP-50, which is the detection results with different thresholds of Intersection over Union (IoU) in the Non-Maximum Suppression (NMS). Furthermore, due to the different object detection methods employing different backbones and detection strategies, we also introduce the inference time ($ms$), Parameter (Million as a unit), and Tera Flops for efficient comparison as shown in  \textit{Table~\ref{tab:efficiency}}. 
%
\subsection{Quantitative Results}
\textbf{Fully supervised multi-object detection.} 
In this paper, we compare 9 most popular object detection methods, including the YOLO-based detection~\cite{yolox2021}, two-stage detection~\cite{ren2016faster}, one-stage detection~\cite{peize2020sparse}, and Transformer-based detection~\cite{carion2020endtoend,zhu2021deformable,meng2021-CondDETR,liu2022dabdetr,li2022grounded,li2022exploring,zhou2022detecting,zong2023detrs,zhang2022dino,Zhang_2023_CVPR}. As shown in Tables~\ref{tab:fully_supervised}, the Relation-DETR~\cite{hou2024relation} obtains the highest detection performance with 85.6\% and 95.6\% mAP, which significantly outperforms the second-best method DDQ~\cite{Zhang_2023_CVPR} with 0.2\% and 0.2\% improvement. For four different mainstream detection methods, Transformer-based detection can significantly outperform others; however, their efficiency is seen in more cumbersome computation and longer inference time when compared to methods based on convolutional neural networks. The method YOLOX\cite{yolox2021} can reach considerable results in detection while only taking $8.942\;ms$ inference time with only $0.047$ TFlops to process one image, which can maintain high detection accuracy with the fastest inference speed. The quantitative data underscores the robustness of these approaches in medical imaging tasks, providing a solid benchmark for future advancements in automated prenatal diagnostics. Detailed comparisons highlight the nuanced strengths of each model, facilitating targeted improvements in subsequent research.

\textbf{Semi-supervised multi-object detection.} 
In the semi-supervised multi-object detection evaluation shown in Table~\ref{tab:semi-supervised}, our results highlight significant performance variations among baseline methods~\cite{liu2022unbiased, chen2022label, wang2023consistent, zhang2023semi, shehzadi2024sparse, pu2025anatomical} across different labeled data percentages (5\% and 10\%) and hospital datasets. The analysis reveals that transformer-based approaches~\cite{zhang2023semi, shehzadi2024sparse} exhibit superior adaptability and robustness, particularly in data with insufficient annotation scenarios. Notably, certain methods demonstrate consistent improvements as the percentage of labeled data increases, indicating their effectiveness in leveraging unlabeled data for enhanced learning. Additionally, the comparison across hospitals underscores the importance of domain adaptation techniques in mitigating dataset biases. These findings suggest that while some methods excel in specific conditions, a comprehensive approach integrating advanced semi-supervised strategies is crucial for achieving optimal detection accuracy and generalization in diverse clinical settings. 

\textbf{UDA and Source-free UDA in multi-object detection.} 
In our experiments of UDA for multi-object detection (see Table~\ref{tab:uda}), we observed distinctive patterns in model performance across various clinical settings. UDA techniques uniquely bridge domain gaps without relying on target domain labels, a capability not fully realized by fully supervised and semi-supervised methods. We have compared six UDA methods proposed for both natural~\cite{li2022sigma,li2023sigma++,cao2023contrastive} and medical~\cite{pu2024m3,pu2024unsupervised} image domains. Our findings reveal that certain UDA strategies effectively mitigate dataset biases, enhancing cross-domain generalization. Specifically, methods incorporating anatomical consistency and structural knowledge excel at transferring learned representations across different hospitals.

As shown in Table~\ref{tab:sfuda}, source-free UDA approaches~\cite{li2022cross,chu2023adversarial,vs2023instance,pu2025leveraging} further extend this adaptability, demonstrating significant promise in scenarios where original datasets are inaccessible. These methods maintain robust performance even when the source data is unavailable, highlighting their practical utility in real-world applications. By leveraging intrinsic features and domain-invariant characteristics, source-free UDA ensures consistent accuracy across diverse medical environments. This adaptability underscores the critical role of tailored UDA methodologies in advancing automated diagnostic tools for early pregnancy ultrasound screening. To summarize, UDA and source-free UDA methods offer a comprehensive solution to the challenges posed by domain shifts. They enhance the reliability and effectiveness of multi-object detection systems, ensuring they can operate seamlessly across varied clinical settings. The integration of these advanced techniques not only improves diagnostic accuracy but also paves the way for more versatile and adaptable medical imaging analysis tools. Thus, the combination of UDA and source-free UDA represents a significant step forward in addressing the complexities of multi-center data in medical applications. 

{\textbf{Failure Mode Analysis.} We highlight that tiny anatomical structures (such as the Nasal Bone) occupy merely 0.2\% to 0.5\% of the total detection area, making early-pregnancy ultrasound uniquely challenging due to Extreme Scale Variation. Furthermore, our analysis reveals that cross-center generalization performance drops are highly correlated with changes in imaging devices. UDA methods incorporating anatomical consistency performed best, suggesting that enforcing structural topology constraints is the most promising future research direction.}

\textbf{Note:} We also conducted the statistical analysis of FUSEP dataset, which can be checked in \emph{{Section Appendix}} (Figures~\ref{fig:avg_area_per_class} and ~\ref{fig:bbox_counts_per_class}).

\begin{table}[t]
\centering
\caption{Computational cost analysis of different detection methods. Times are reported in milliseconds (ms), parameters in millions (M), and training times in minutes (min).}
\label{tab:efficiency}
\scriptsize 
\setlength{\tabcolsep}{2.5pt} 
\begin{tabular}{llccccccc}
\toprule
\textbf{Method} & \textbf{Backbone} & \makecell{\textbf{Infer.}\\ \textbf{Time}\\ \textbf{(ms)} $\downarrow$} & \makecell{\textbf{Params}\\ \textbf{(M)} $\downarrow$} & \makecell{\textbf{TFlops}\\ $\downarrow$} & \makecell{\textbf{Train/}\\ \textbf{Epoch}\\ \textbf{(min)} $\downarrow$} & \makecell{\textbf{Conv.}\\ \textbf{Epochs}\\ $\downarrow$} & \makecell{\textbf{GPU}\\ \textbf{Hrs}\\ $\downarrow$} & \makecell{\textbf{Mem}\\ \textbf{(GB)}\\ $\downarrow$} \\
\midrule
Faster R-CNN & ResNet-50 & 41.405 & 41.1 & 0.192 & 4.5 & 24 & 1.8 & 4.6 \\
DETR & ResNet-50 & 41.558 & 48.5 & 0.087 & 48.2 & 33 & 5.4 & 6.4 \\
Deformable-DETR & ResNet-50 & 40.101 & 65.2 & 0.180 & 8.0 & 21 & 4.6 & 6.9 \\
YOLOX & YOLOX-L & 8.942 & 27.6 & 0.047 & 4.1 & 13 & 1.6 & 3.2 \\
ViTDet & ViT-B & 113.664 & 150.1 & 0.840 & 18.4 & 42 & 7.4 & 20.3 \\
CO-DETR & ResNet-50 & 46.133 & 88.5 & 0.244 & 7.4 & 19 & 4.1 & 7.1 \\
DINO & ResNet-50 & 47.563 & 90.6 & 0.256 & 5.7 & 28 & 2.6 & 7.5 \\
DDQ & ResNet-50 & 48.348 & 111.2 & 0.258 & 4.9 & 22 & 2.0 & 8.0 \\
Relation-DETR & ResNet-50 & 47.315 & 91.5 & 0.264 & 5.8 & 16 & 2.3 & 6.7 \\
\bottomrule
\end{tabular}
\end{table}

\section{Conclusion} 
We introduce FUSEP, the first open-source ultrasound dataset to date designed for real-world early-pregnancy fetal screening. Collected from three hospitals, FUSEP provides multi-center clinical data with strong practical relevance and diversity. Based on this dataset, we systematically investigate the detection of 14 densely distributed fetal anatomical structures across two important ultrasound views, and evaluate a wide range of advanced detection methods for this challenging task. Our experiments demonstrate the effectiveness and value of FUSEP as a new benchmark for early-pregnancy fetal ultrasound analysis. By providing both a publicly available dataset and standardized evaluation settings, FUSEP can facilitate the development of more accurate and robust medical multi-structure detection techniques. We believe that FUSEP will better support intelligent fetal ultrasound screening and provide a valuable resource for the broader ultrasound research community.

\textbf{Future Work.} Our study represents a significant step forward in intelligent ultrasound screening of fetuses in early pregnancy, but there is still significant potential for improvement.
Key directions for the future include: 1) Extending both our data samples and the collection of more types of fetal views to cover the entirety of early pregnancy.
2) The collection of data from different hospital centers has been further evaluated, which is essential for real clinical practice.
3) Labeling segmentation masks of anatomical structures, which are essential for the measurement and estimation of fetal growth parameters in early pregnancy, facilitates the development of assisted screening and diagnostic systems for early fetal pregnancy.

\section*{Acknowledgments} 
This work was supported in part by the National Natural Science Foundation of China under Grant 62227808.

\clearpage

\bibliographystyle{ACM-Reference-Format}
\bibliography{Reference}

\appendix

\renewcommand\thefigure{A2}
\begin{figure*}[t!]
    \centering
    \includegraphics[width=0.999\textwidth]{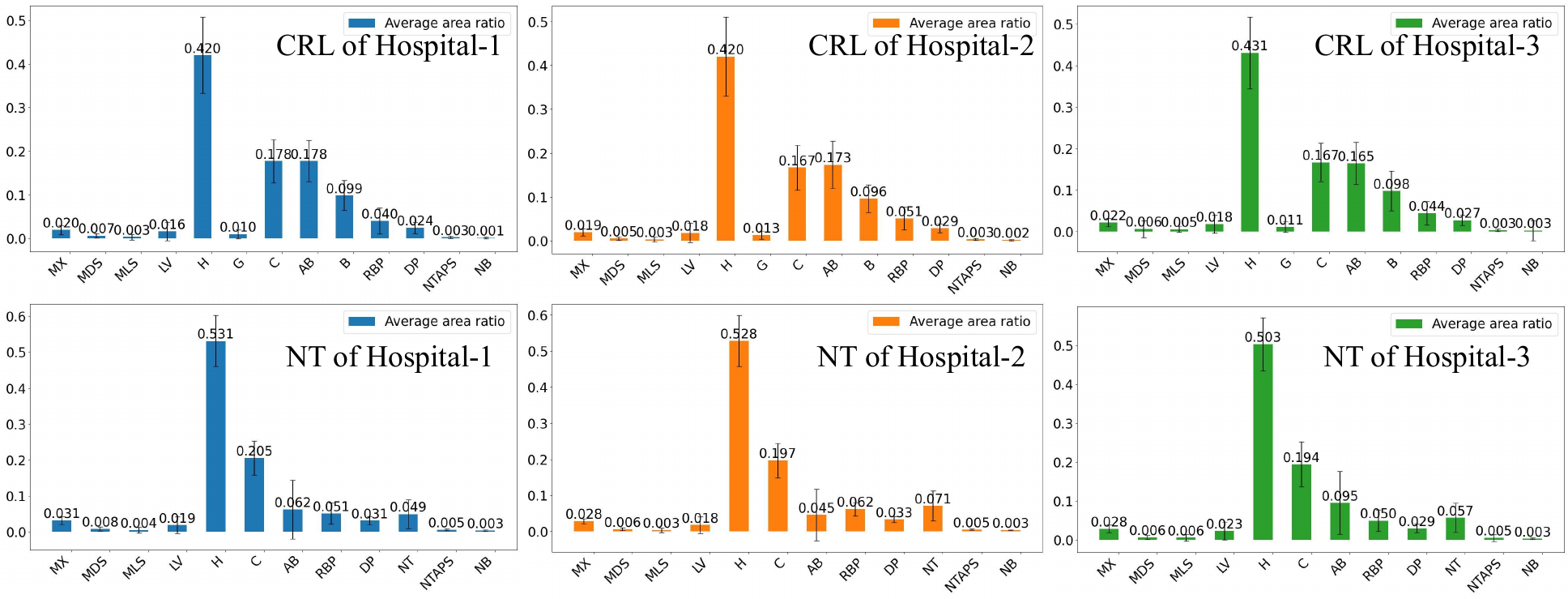}
    \caption{The overall area ratio of different structures in CRL and NT views.}
    \label{fig:avg_area_per_class}
\end{figure*}
\renewcommand\thefigure{A3}
\begin{figure}
    \centering
    \includegraphics[width=0.49\textwidth]{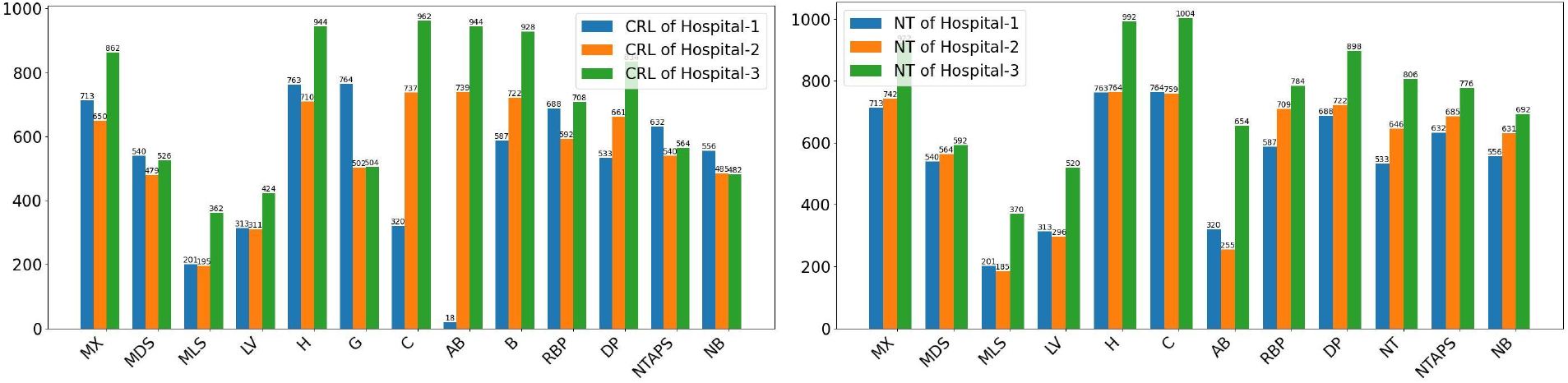}
    \caption{The counts of bounding boxes per class, the CRL and NT views are presented in the left and right bar charts, respectively. The bars with different colours represent the number of bounding boxes in the overall training, validation, and test sets for each class.}
\label{fig:bbox_counts_per_class}
\end{figure}

\section*{Appendix}

\subsection*{A. Statistical Analysis of \textbf{FUSEP} Dataset}
\textbf{\textit{The Annotations Analysis of FUSEP Dataset.}} Figure~\ref{fig:avg_area_per_class} shows the area ratios of each object, which indicate the difficulty of the detection. (The objects with larger areas are easier to detect). Figure~\ref{fig:bbox_counts_per_class} shows the total object quantity annotated in datasets, indicating that not all detection targets are fully included. This is because the scanned fetuses are at different stages of development, and some organs have not yet differentiated or developed into shape.
\renewcommand\thefigure{A1}
\begin{figure}[htbp]
    \centering
    \includegraphics[width=0.49\textwidth]{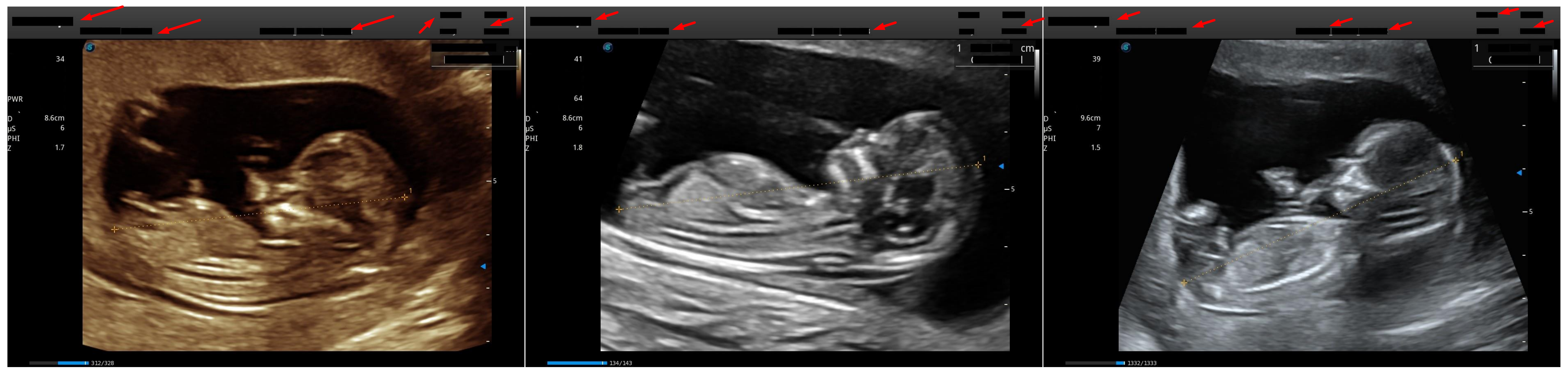}
    \caption{The example of ultrasound images. The arrows indicate that the personal information that may be contained in our ultrasound images has been covered.}
    \label{appendix:fig_1}
\end{figure}

For detailed analysis, Figure~\ref{fig:bbox_counts_per_class} demonstrates that the G, LV and NB are the least annotated among all detection categories, which are only 1,989, 1,038 and 1,314 in overall 1,816 samples of CRL view. For the NT view, the MLS and AB are significantly less than other categories, with only 756 and 1879 annotations of 2,201 samples. The main reason for this is that early pregnancy does not require the detection of those objects. For example, the structures LV of the CRL view, as well as AB and MLS of the NT view, are not necessary for early pregnancy anatomy detection. Since there are some non-standard scanned images in our dataset, which often lead to errors in the measurement of fetal growth parameters. Hence, we also annotate those structures that do not need to be observed in the CRL and NT views for standard view recognition. {This severe class imbalance and Feature Sparsity directly cause these specific classes to achieve lower AP scores in the baseline evaluations.}

Moreover, Figure~\ref{fig:avg_area_per_class} illustrates that the bounding boxes of different objects have a significant difference in the difficulty of detection. The top row of Figure~\ref{fig:avg_area_per_class} shows the average ratio and the standard deviation of all bounding boxes within a single image in CRL view; the H have the largest area that occupies 40\% to 50\% of the total detection area. In contrast, the NB, NTAPS, MDS, and G are the smallest objects that only occupy the total area ratio of 0.002, 0.004, 0.010 and 0.012, showcasing more challenging detection of these four objects. In the bottom row of Figure~\ref{fig:avg_area_per_class}, structure H further increases the area ratio to 50\%-60\% in NT view, while the structures NB, NTAPS, MDS and MLS are the challenging objects with only a ratio of total detection areas in 0.003, 0.005, 0.006 and 0.003, respectively. {This Extreme Scale Variation makes detection uniquely challenging compared to natural images.}

This study also denotes that our proposed \textbf{FUSEP} dataset is non-biased and capable of being applied and deployed in real-world medical scenarios. We provide all device information for each sample, which allows future research not only in object detection problems but also further in domain adaptation tasks. Demonstrates the great potential of our dataset and its significant contribution to the healthcare community.

\end{document}